\documentclass{article}

\usepackage{arxiv}

\usepackage{algorithm} 
\usepackage{algorithmic} 

\usepackage{bbm}
\usepackage{subcaption}

\usepackage[utf8]{inputenc} 
\usepackage[T1]{fontenc}    
\usepackage{hyperref}       
\usepackage{url}            
\usepackage{booktabs}       
\usepackage{amsfonts}       
\usepackage{nicefrac}       
\usepackage{microtype}      
\usepackage{lipsum}
\usepackage{graphicx}
\graphicspath{ {./images/} }

\title{Conspiracy in the Time of Corona: Automatic detection of Covid-19 Conspiracy Theories in Social Media and the News}

\author{
 Shadi Shahsavari\\
  Electrical and Computer Engineering\\
  UCLA\\
  Los Angeles, CA, USA
  \texttt{shadihpp@g.ucla.edu} \\
  \And
Pavan Holur\\
  Electrical and Computer Engineering\\
  UCLA\\
  Los Angeles, CA, USA
  \texttt{pholur@g.ucla.edu} \\
  \And
 Timothy R. Tangherlini\\
 Scandinavian Section\\
  UCLA\\
  Los Angeles, CA, USA
  \texttt{tango@g.ucla.edu} \\
  \And
Vwani Roychowdhury\\
  Electrical and Computer Engineering\\
  UCLA\\
  Los Angeles, CA, USA
  \texttt{vwani@g.ucla.edu} \\
}

\begin{document}
\maketitle
\begin{abstract}
Rumors and conspiracy theories thrive in environments of low confidence and low trust. Consequently, it is not surprising that Covid-19 related rumors and conspiracy theories are proliferating. This phenomenon is particularly acute given the lack of any authoritative scientific consensus on the virus, its spread and containment, or on the long term social and economic ramifications of the pandemic. Among the stories currently circulating on social media and in the news are ones suggesting that the 5G network activates the virus or causes physiological change that can kill, that the pandemic is a hoax perpetrated by a global cabal, that the virus is a bio-weapon released deliberately by the Chinese, or that Bill Gates is using the pandemic to launch a global surveillance regime under cover of vaccination. While some may be quick to dismiss these stories as having little impact on real-world behavior, recent events including the destruction of property, racially fueled attacks against Asian Americans, and demonstrations espousing resistance to public health orders countermand such conclusions. Rumors and conspiracy theories about the pandemic pose a significant threat not only to democratic institutions such as a free, open and trusted press, but also to the physical well-being of the citizenry. Inspired by narrative theory, we crawl social media sites and news reports and, through the application of automated machine-learning methods, discover the underlying narrative frameworks supporting the generation of these stories. We show how the various narrative frameworks fueling rumors and conspiracy theories rely on the alignment of otherwise disparate domains of knowledge, and consider how they attach to the broader reporting on the pandemic. These alignments and attachments, which can be monitored in near real-time, may be useful for identifying areas in the news that are particularly vulnerable to reinterpretation by conspiracy theorists. Understanding the dynamics of storytelling on social media and the narrative frameworks that provide the generative basis for these stories may also be helpful for devising methods to disrupt their spread. 
\end{abstract}


\section{Introduction}

As the Covid-19 pandemic continues its unrelenting global march, stories about its origins, possible cures, and appropriate responses are tearing through social media and dominating the news cycle. While many of the stories in the news media are the product of fact-based reporting, many of those circulating on social media are anecdotal and the product of speculation, wishful thinking, or conspiratorial fantasy. Given the lack of a strong scientific and governmental consensus on how to combat the virus, people are turning to informal information sources such as social media to share their thoughts and experiences, and to discuss possible responses. At the same time, the news media is reporting on the actions individuals and groups are taking across the globe, including ingesting home remedies or defying stay at home orders, and on the information motivating those actions.\footnote{The broad scale use of hydroxychloroquine as a possible treatment for the virus, which first gained a foothold in social media, is only one example of this feedback loop in action \cite{samuels2020hydroxy}. Numerous other claims, despite being repeatedly shown to be false, persist in part because of this close connection between social media and the news. \cite{blake2020zombie}} Consequently, news and social media have become closely intertwined, with informal and potentially misleading stories entangled with fact-based reporting: social media posts back up claims with links to news stories, while the news reports on stories trending on social media. To complicate matters, not all sites purporting to be news media are reputable, while reputable sites have reported unsubstantiated or inaccurate information. Because of the very high volume of information circulating on and across these platforms, and the speed at which new information enters this information ecosystem, fact-checking organizations have been overwhelmed. The chief operating officer of Snopes, for example, has pointed out that, “[there] are rumors and grifts and scams that are causing real catastrophic consequences for people at risk… It’s the deadliest information crisis we might ever possibly have,” and notes that his group and others like it are “grossly unprepared” \cite{leskin2020snopes}.

Devising computational methods for disentangling misleading stories from the actual news is a pressing need. Such methods could be used to support fact checking organizations, and help identify and deter the spread of misleading stories; ultimately, they may help prevent people from making potentially catastrophic decisions, such as self-medicating with chloroquine phosphate, bleach or alcohol, or resisting efforts at containment that require participation by an entire citizenry.\footnote{An Arizona man died from using an aquarium additive that contained chloroquine phosphate as a source of chloroquine, a potential “miracle cure” touted by various sources, including the president\cite{waldrop2020arizona}. Several poison control centers had to make press releases warning people not to drink or gargle with bleach\cite{capatides2020corona}. The governor of Nairobi included small bottles of cognac in the Covid-19 care kits distributed to citizens, erroneously indicating that WHO considers alcohol a “throat sanitizer”\cite{feleke2020kenya}.} 

As decades of research into folklore has shown, stories such as those circulating on social media, however anecdotal, are not created from whole cloth, but rely on existing stories, story structures, and conceptual frameworks that inform the world view of individuals and their broader cultural groups \cite{vonsydow1948tradition}\cite{pentikainen1976takalo}\cite{siikala1990oral}\cite{tangherlini2018genmodel}. Taken together, these three features (a shared world view, a reservoir of existing stories, and a shared understanding of story structure) allow people to easily generate stories acceptable to their group, and to try to convince others to see the world as they do. 

Inspired by the narratological work of Algirdas Greimas \cite{greimas1966elements}, and the social discourse work of Joshua Waletzky and William Labov \cite{labov1967narrative}, we devise an automated pipeline that determines the narrative frameworks undergirding knowledge domains, in this case those related to the Covid-19 pandemic.\cite{tangherlini_plos} We also borrow from George Boole's famous definition of a domain of discourse, recognizing that in any such domain there are informal and constantly negotiated limits on what can be said \cite{boole1854discourse}. We conceptualize a narrative framework as a network comprising the actants (people, organizations, places and things) and their interactant relationships that are activated in any storytelling related to the pandemic, be it a journalistic account or an informal anecdote \cite{tangherlini2016mommy}\cite{tangherlini_plos}. In this model of story telling, individuals activate a small subset of the available actants and interactant relationships that exist in a discourse domain. 

To determine the extent of narrative material available---the actants and their complex, content dependent interactant relationships---we aggregate all the posts or reports from a social media platform or news aggregator. For social media in particular, we recognize that participants in an online conversation rarely recount the entire scope of a story, choosing instead to tell parts of it \cite{laudun2001talk}. Yet even partial stories activate some small group of actants and relationships available in the broader discourse. We conceptualize this as a weighting of a subgraph of the larger narrative framework network.

By applying the narrative framework discovery pipeline to tens of thousands of social media posts and news stories, all centered on conspiracy theories related to the Covid-19 pandemic, we uncover five central phenomena: (i) the attempt to incorporate the pandemic into well-known conspiracy theories, such as Q-Anon; (ii) the emergence of new conspiracy theories, such as one suggesting that the 5G cellular network is the root cause of the virus; (iii) the alignment of various conspiracy theories to form larger ones, such as one suggesting that Bill Gates is using the virus as a cover for his desire to create a worldwide surveillance state through the enforcement of a global vaccination program, thereby aligning it with  anti-vaxx conspiracy theories; (iv) the nucleation of potential conspiracy theories that may grow into a larger theory or be subsumed in one of the existing or emerging theories; and (v) the interaction of these conspiracy theories with the news, where certain factual events, such as the setting up of tents in Central Park for a field hospital, are linked to the Q-Anon conspiracy theory that includes aspects of child sex-trafficking, underground tunnels, and Hollywood stars best known in recent years from the Pizzagate conspiracy theory. 

Running the pipeline on a daily basis allows us to capture snapshots of the dynamics of the entanglement of news and social media, revealing ebbs and flows in the overall story graph, while highlighting the parts of the news graph that are susceptible to being linked to rumors and conspiratorial thinking. 

\section{Prior work}
Conspiracy theories (along with rumors and other stories told as true) circulate rapidly when access to trustworthy information is low, when trust in accessible information and its sources is low, when high quality information is hard to come by, or a combination of these factors \cite{allport1947psych}\cite{rosnow1980psych}\cite{shibutani1966rumor}\cite{rosnow1976rumor}\cite{campion2017rumor}\cite{lewandowski2020conspiracy}\cite{bessi2015conspiracy}. In these situations, people exchange informal stories about what they believe is happening, and negotiate possible actions and reactions, even as that event unfolds around them. Research into the circulation of highly believable stories in the context of natural disasters such as Hurricane Katrina \cite{lindahl2012katrina}, man-made crises such as the 9/11 terrorist attacks in 2001 \cite{fine2013global} and the Boston Marathon bombings in 2013 \cite{starbird2014boston}, or crises with longer time horizons such as disease \cite{bandari2017resistant} \cite{heller2015rumors} \cite{kitta2012vaccination}, has confirmed the explanatory role storytelling plays in these events, while underscoring the impact that these stories, including incomplete ones, can have on solidifying beliefs and inspiring real-world action \cite{kuperman2004genocide}\cite{fine2001whispers}. 

The goal of telling stories in these situations is at least in part to reach group-wide consensus on the causes of the threat or disruption, the possible strategies that are appropriate to counteract that threat, and the likely outcomes of a particular strategy \cite{tangherlini2018genmodel}. In an unfolding crisis, stories often provide a likely cause or origin for the threat, and propose possible strategies for counteracting that threat; the implementation of those strategies can move into real-world action, with the strategy and results playing themselves out in the physical world. This pattern has repeated itself many times throughout history, including during recent events such as Edgar Welch’s attempt to “free” children allegedly being held in a Washington DC pizza parlor \cite{metaxas2017infamous}, the genocidal eruptions that crippled Rwanda with paroxysms of violence in 1994 \cite{desforges1999genocide}, and the global anti-vaccination movement that continues to threaten global health \cite{capurro2018measles}.

Although the hyperactive transmission of rumors often subsides once credible and authoritative information is brought to the forefront \cite{victor1993satanic} \cite{bandari2017resistant}, the underlying narrative frameworks that act as a generative reservoir for these stories do not disappear. Even in times of relative calm where people have high trust in their information sources and high confidence in the information being disseminated through those sources, stories based on the underlying narrative frameworks continue to be told, and remain circulating with much lower frequency in and across various social groups. This endemic reservoir of narrative frameworks serves multiple cultural functions. It supports the enculturation of new members, proffering a dynamic environment for the ongoing negotiation of the group’s underlying cultural ideology. Also, it provides a ready store of explanatory communal knowledge about potential threats and disruptions—--their origins and their particular targets-—-as well as a repertoire of potentially successful strategies for dealing with those threats \cite{decerteau1984practice}. When something does happen that needs explanation--—and a possible response—--but for which high trust or high confidence information sources do not exist, the story generation mechanism can shift into high gear. 

The endemic reservoir of narrative frameworks that exists in any population is not immutable. Indeed, it is the ability of people to change and shape their stories to fit the specific information and explanatory needs of their social groups that makes them particularly fit for rapid and broad circulation in and across groups \cite{tangherlini2018genmodel}. While the stability in a story telling tradition suggests that the actants and their relationships are slow to change, their constant activation through the process of storytelling leads to dynamic changes in the weighting of those narrative framework networks; new actants and relationships can be added and, if they become the subject of frequent storytelling, can become increasingly central to the tradition.  

Because of their explanatory power, stories can be linked into cycles to form conspiracy theories, often bringing together normally disparate domains of human interaction into a single, explanatory realm \cite{lewandowski2020conspiracy} \cite{knight2003conspiracies}. Although a conspiracy theory may not ever be recounted in its entirety, members of the communities in which such a theory circulates have, through repeated interactions, internalized the “immanent” narrative that comprises the overall conspiracy theory \cite{clover1986prose}. In turn, conspiracy theories can be linked to provide a hermetic and totalizing world view redolent of monological thinking \cite{goertzel1994belief}, and can thereby provide explanations for otherwise seemingly disjoint events while aligning possible strategies for dealing with the event to the teller’s cultural ideology \cite{tangherlini2018genmodel}. 

Summarizing the storytelling of thousands of story tellers and presenting these stories in an organized fashion has been an open problem in folkloristics since the beginning of the field. The representation of narratives as network graphs has been a desiderata in narrative studies at least since the formalist studies of Vladimir Propp \cite{propp1928morphology}. Lehnert, in her work on the representation of complex narratives as graphs, notes that these representations have the ability to “reveal which concepts are central to the story” \cite{lehnert1980narrative}. In other work specifically focused on informal storytelling, Bearman and Stovel point out that, “By representing complex event sequences as networks, we are easily able to observe and measure structural features of narratives that may otherwise be difficult to see” \cite{bearman2000becoming}. Later work on diverse corpora including national security documents has shown the applicability of the approach to a broad range of data resources \cite{mohr2013graphing}. 

The automatic extraction of these graphs, however, has been elusive given the computational challenges inherent in the problem. In the context of conspiracy theories, preliminary work has successfully shown how simple S-V-O extractions can be matched to a broader topic model of a large corpus of conspiracy theories \cite{samory2018conspiracy}. Work in our group has shown how the extraction of more complex structures and the alignment into complex narrative graphs provide a clear overview of the narrative frameworks supporting the decision to seek exemptions from vaccination among antivax communities posting on parenting blogs \cite{tangherlini_plos}. 

Recent work on rumors and conspiracies focus specifically on the Covid-19 pandemic \cite{ferrara2020twitbots}. An analysis of German Facebook groups whose discussions center on the pandemic uses a similar named entity analysis to our methods, and shows a strong tendency among the Facebook group members to resist the news reported by recognized journalistic sources \cite{boberg2020populism}. An examination of 4Chan that employs network analysis techniques and entity rankings traces the emergence of sino-phobic attitudes on social media, which are echoed in our narrative frameworks \cite{schild2020sinophobe}.

In forthcoming work, we have shown how conspiracy theories align disparate domains of human knowledge or interaction through the interpretation of various types of information not broadly accessible outside the community of conspiracy theorists \cite{tangherlini_plos}. We have also shown that conspiracy theories, like rumors and legends on which they are based, are opportunistic, taking advantage of low information environments to align the conspiracy theory to unexplained events in the actual news \cite{bandari2017resistant}. Such an alignment provides an overarching explanation for otherwise inexplicable events, and fits neatly into the world view of the conspiracy theorists. The explanatory power of this storytelling can also entice new members to the group, ultimately getting them to ascribe to the worldview of that group.

\section{Data}
Data for this study were derived from two main sources, one a concatenation of social media resources composed largely of forum discussions, and the other a concatenation of Covid-19 related news reports from reputable journalistic sources. 

We devised a scraper to collect publicly available data from Reddit subreddits and from 4Chan threads related to the pandemic. The subreddits and threads were evaluated by three independent evaluators, and selected only if there was consensus on their relevance. All of the data are available in our Open Science Framework data repository \cite{shahsavari2020data}.\footnote{We ensured that our data was free from PII, and that our data collection was allowed by the terms of service of the two sites. To the best of our knowledge, neither our corpus nor the news data contains data from private discussions, private chat rooms, or any other sources with restrictions on access for public use.}

For 4chan, we collected $\sim 200$ links to threads for the term ``coronavirus'' on 4chan.org, resulting in a corpus of $14712$ posts. The first post in our corpus was published on March 28, 2020 and the final post was published on April 17, 2020. For Reddit, we accessed $\sim 100$ threads on various subreddits with $4377$ posts scraped from the top comments. Specifically, we targeted r/coronavirus and r/covid19, along with threads from r/conspiracy concentrating on Corona virus. We removed images, URLs, advertisements, and non-English text strings from both sources to create our research corpus. After running our pipeline, we were able to extract $87079$ relationships from these social media posts.

For news reports, we relied on Google’s GDELT project, an Open Source platform that scrapes web news (in addition to print and broadcast) from around the world (https://www.gdeltproject.org/).\footnote{Research use of the platform is explicitly permitted on the GDELT ``about'' pages.} Our search constraints through this dynamic corpus of news reports included a first-order search for {conspiracy} theories and a second-order search for {coronavirus}. The filtering constraints included the origin of the news source (United States) and the language of publication (English). The top $100$ news articles (as sorted by the GDELT engine) were aggregated daily from January 1, 2020 to April 14, 2020  (prior to filtering), and the body of each filtered news report was scraped (with Newspaper3K), cleaned and staged for our pipeline to extract sentence-level relationships between key actors. We extracted $\sim 60$ relationships from each report, $\sim 50$ filtered news reports per day and $324510$ relationships. 

\section{Methods}

We estimate narrative networks that represent the underlying structure of conspiracy theories in the social media corpus (4Chan, Reddit) where they are most likely to originate, and the corresponding reporting about them in the news corpus (GDELT). This allows us 
to analyze the interplay between the two corpora and to track the time-correlation and pervasive flow of information from one corpus to the other. The following subsections introduce the graphical narrative model applied on social media and the pipeline to process the news reports.

\subsection{A Graphical Narrative model and its estimation}
We model narratives as being generated by an underlying graphical model \cite{tangherlini_plos}. 
Such a narrative graph is characterized by a set of $n$ nodes that represent the actants, a set  $r$ relationships $R=\{R_1,R_2,\dots,R_r\}$ that define the edges,  and $k$ contexts, $C=\{C_1,C_2,\dots,C_k\}$, which provides a hierarchical structure to the network. These parameters are either given \textit{a priori} or estimated from the data. A context $C_i$ is a hidden parameter or the `phase' of the underlying system which defines the particular environment in which actants operate. It expresses itself in the distributions of the relationships among the actants, and is captured by a labeled and weighted network $G_{C_i}(V_{C_i},E_{C_i})$. Here, $V_{C_i}=\{A_1,A_2,\dots,A_n\}$, where each $A_j$ is an actant.
The edge set $E_{C_i}$ consists of $m_{C_i}$ ordered pairs $e_{C_i,j}=(A_{j_1},A_{j_2})$, where each such pair is labeled with a distribution over the relationship set $R$.

Each post to a thread describes relationships among a subset of actants. A user picks a context $C_i$ and then draws a set of actants and edges between them from the network $G_{C_i}(V_{C_i},E_{C_i})$. The output of this generative process are the posts to the forums. From a machine learning perspective, given this text data, we need to estimate all the hidden parameters of the model: the actants, the contexts, the set of relationships, and the edges and their labels. In other words, it is necessary to jointly estimate all the parameters of the different layers of the model. We summarize this process below, while the code is available on our github repository.\footnote{https://tinyurl.com/ydftuez6}

First, each sentence in our corpus is processed to extract various patterns of syntax relationship tuples. These tuples are described as $(arg_1,rel,arg_2)$ where each $arg_i$ is a noun phrase and $rel$ is a verb or other type of phrase. Our relations extraction framework combines dependency parse tree and Semantic Role Labeling (SRL) \cite{collobert2011natural} tools used in Natural Language Processing (NLP). We design sets of relationship patterns that are important in articulating narratives, and extract them from the parsed outputs of the NLP models. 

The noun phrases $arg_1$ and $arg_2$ along with the phrases obtained using a Named Entity Recognition tool \cite{akbik2019flair} provide information about the actant nodes and their different mentions in the graphical model. These phrases need to be contextually aggregated across the entire corpus to group them into semantic categories or actants. Several different methods of clustering noun phrases and entity mentions into semantically meaningful groups have been proposed in the literature \cite{Clustype}. 

In this paper, we follow a recently introduced method where the noun phrases are aggregated into contextually relevant groups referred to as super-nodes or Contextual Groups (CGs). To achieve this aggregation, we generate a list of names across the corpus by applying an NER algorithm. This list of named entities then acts as a seed list to start the process of creating CGs from the noun phrases. In particular, we group nouns in the NER list using co-occurrence information from phrases; for example, ``corona" and ``virus" tend to appear in phrases together.  Table \ref{tab:contextgroups} shows examples of CGs. In prior work \cite{tangherlini_plos}, we have shown that the final network comprising sub-nodes (see below) and their relationship edges is not sensitive to the exact size and content of the contextual groups derived at this stage. This CG grouping, which we apply prior to applying k-Means clustering on word embedding (see below), enables us to distill the noun phrases into groups of phrases that have a semantic bias. This distillation mitigates the inherent noise issues with word embeddings when they are directly used to cluster heterogeneous sets of words over a large corpus \cite{samory2018spies}. The CGs can also be viewed as defining macro-contexts in the underlying generative narrative framework. 

Next, we leverage recent work on word embedding that allows for the capture of both word and phrase semantics. In particular, the noun phrases in the same contextual group are clustered using an unsupervised k-Means clustering algorithm according to their embeddings derived from a well-known word embedding tool BERT \cite{devlin2018bert}. Note that the sub-nodes are not disjoint in their noun-phrase contents: the same phrase might appear in different sub-nodes. The sub-nodes, when viewed as groups, act as different micro-contexts in the underlying model. 

The final automatically derived narrative framework graph is composed of phrase-cluster nodes that we also refer to as sub-nodes. We automatically label these sub-nodes based on the TFIDF scores of the words in each cluster. The edges among the sub-nodes are obtained by aggregating the relationships among all pairs of noun phrases. Each edge thus has a set of relationship phrases associated with it, and the number of relationships can serve as the weight of an edge. The relationship set defining an edge can be further partitioned  into semantically homogeneous groups by clustering their BERT embeddings \cite{devlin2018bert}.  

\subsection{Communities in Narrative Networks and their estimations}

Conspiracy theories tend to connect different existing domains of human activities via creative speculation or access to ``hidden knowledge'' to create an all encompassing narrative. We thus expect the narrative networks that we construct to have clusters of nodes and edges that focus on different domains. These domain-related clusters will be densely connected within themselves, with a sparser set of edges connecting them with others. Such densely connected components of networks are called ``communities''. Traversing these clusters can create a complete narrative framework, while storytelling often activates some small subset of subnodes and edges. 

Given the unsettled nature of discussions concerning the Covid-19 pandemic, posts in social media in the aggregate likely articulate  multiple competing conspiracy theories (which we do not know prior to our discovering them). Consequently, one would expect to find a number of communities in the overall network, with narrative frameworks comprised of a selection of subnodes from various communities, perhaps with a large number of subnodes from a smaller number of the activated communities. 



In order to capture any such hierarchical structures in our narrative networks,  we computed overlapping community structures, where each community is defined by (i) a core set of nodes that constitute its backbone, and (ii) a set peripheral nodes of varying significance, which it shares with other communities.  This is done by running the Louvain \cite{blondel2008louvain} greedy community detection algorithms multiple times and then grouping nodes that co-occur above a certain threshold value to form the cores of different communities. Overlapping nodes are then brought into a community by relaxing this threshold. More information about this approach can be found in previous work \cite{shahsavari2020goodreads}.

\begin{table}[]
    \centering
    \begin{tabular}{|c|c|}
    \hline
       \textbf{Contextual group}  & \textbf{Frequency}  \\\hline \hline
       {virus, viruses, corona}&2966\\\hline
       {china}&7865\\\hline
       {chinese, chineses, government}&3622\\\hline
       {wuhan, lab}&2691\\\hline
       {trumps, trump, donald}&1913\\\hline
       {gate, gates, bill}&1158\\\hline
       {jews, jew}&572\\\hline
       {usa, military}&525\\\hline
       {doctor, doctors, nurse}&520\\\hline
    \end{tabular}
    \caption{Examples of contextual groups}
    \label{tab:contextgroups}
\end{table}





\subsection{Baseline News Reports about Conspiracy Theories}
To aggregate the published news, we consider ($1$-day time-shifted) intervals of $5$ days. This sliding window builds $s = 101$ segments from January 1, 2020 to April 14, 2020. We have discovered that a longer interval, such as the one chosen here, provides a richer backdrop of actants and their interactions. In addition, \textit{narratives on consecutive days retain much of the larger context, highlighting the context-dependent emergence of new theories and key actants}. 


As described below we used the major actants and their mentions discovered in the social media data to filter the named entities that occur in the news corpus. A co-occurrence network of key actants in news reports (conditioned on those discovered from social media), provides a day-to-day dynamic view of the emergence of various conspiracy theories through time. In addition, we model the flow of information between social media and news reports by monitoring the frequency of occurrence of social media communities (as captured by a group of  representative actants in each community) in the text of news reports.

\subsubsection{Entity Selection using TF-IDF Filtering from the summarized 4Chan and Reddit entities}

Entities extracted from social media provide a rich vocabulary ($v = 11015$ words) to analyze the parallel emergence of conspiracy theories in the news. To highlight the distinct aspects of each 5-day aggregated set of news reports, we compute the TF-IDF vectors across the $s$ segments. The top $25$ actants identified for a particular segment $i$ from the $i^{th}$ row of the $s \times v$ ($i < s$) matrix of TF-IDF values, are merged with the top $100$ highest frequency actants. While the former set provides diverse actants for each segment $i < s$, the latter constant set encourages a richer context to situate these key entities. 

With minimal supervision, a few actant mentions are grouped together including, [trump, donald] : \textbf{donald trump}, [coronavirus, covid19, virus] : \textbf{coronavirus} and [alex, jones] : \textbf{alex jones}. While such groupings are not strictly required and can be done much more systematically (see \cite{shahsavari2020goodreads}), actant-grouping enhances the co-occurrence graph by reducing the sparsity of the adjacency matrix representing subject-object interaction. Let the final set of entities for a segment $i$ be $E_i$.


\subsubsection{Co-occurrence Actant Network Generation}
For each 5-day segment of aggregated news reports, the corpus of extracted relationships $R_i$ and the associated set of entities $E_i$ are parsed with Algorithm~\ref{alg:co-occurrent}. Day-to-day network dynamics describe the occurrence of entities that feature in social media.

\begin{algorithm}
\caption{Co-occurrence Actant Network Generation for a Segment $i < s$ of News}
\label{alg:co-occurrent}
\begin{algorithmic}
\REQUIRE $R_i$ relationship tuples, $E_i$ entities 
\ENSURE $G_i(R_i, E_i)$
\STATE $M\leftarrow[]$
\FOR{$(arg_1,rel,arg_2) \in R_i$}
\STATE $s \leftarrow H(arg_1)$ \COMMENT{$H(arg)$ is the headword of $arg$}
\STATE $o \leftarrow H(arg_2)$
\STATE $r \leftarrow H(rel)$
\IF{($s,o \in E_i$) \,AND \, ($s \neq o$) \, AND \, ($r$ NOT stop word)}
\STATE $M[s,o] \leftarrow M[s,o] + 1$
\STATE $M[o,s] \leftarrow M[o,s] + 1$
\ENDIF
\ENDFOR
\STATE $M_{norm} = normalize(M)$ \COMMENT{along each row}
\STATE $G_i(R_i, E_i) \leftarrow M$ \COMMENT{Color-coded based on the labels of actants decided by the Entity Extractor}
\end{algorithmic}
\end{algorithm}


\subsection{Evaluation}
The proximity between two actants ($a_1$,$a_2$) in a co-occurrence network $G$ may be measured with the Number of Common Neighbors between them. If the adjacent vertices of $a_1$ are $S_{a_1}$ and of $a_2$ are $S_{a_2}$, the score is defined as:
\begin{equation}
\label{eqn:NCN}
    n_{a_1,a_2} = |S_{a_1} \cap S_{a_2}|.
\end{equation}


We detect the presence of communities sourced from social media in news reports across time by computing a Coverage Score $m$. Between community $g_i$ with vertices $V(g_i)$ and text comprised of a set of words $C_{t_i,t_f}$ (where $t_{min} \leq t_i < t_f \leq t_{max}$), where $C$ is a time-limited sub-corpus of either news reports or 4Chan social media, the score is defined as:
\begin{equation}
\label{eqn:CS}
    m_{g_i,C_{t_i,t_j}} = \frac{\sum_{w_g \in V(g_i)}\sum_{w_C \in C_{t_i,t_j}} \mathbbm{1}{(w_g = w_C)}}{|V(g_i)||C_{t_i,t_f}|}.
\end{equation}

We finally compute the ratio of the coverage score for a community $g_i$ against a random community $g*$ obtained by sampling the key entities from social media. Our random communities consist of $20$ samples of $500$ words each. The cross-correlation of this time-stamped score between social media and the news reports provide a monitoring tool for information flow between the corpora.
Our metrics are standard measurements used for clustering evaluations based on ground truth class labels \cite{sklearn_link}.

\begin{algorithm}
\caption{Unsupervised evaluation of communities}
\label{alg:algo_comm}
\begin{algorithmic}
\REQUIRE $C_{i,t}$ News community indexed $i$ at time $t$, $K_j$ Social media community number $j$ 
\ENSURE $Pr_t$ Percentage of coverage for time $t$, $h_t$ Homogeneity at time $t$, $c_t$ Completeness at time $t$, $v_t$ V-Measure at time $t$ 
\STATE $Y_{gr}\leftarrow[]$
\STATE $Y_{pred}\leftarrow[]$
\STATE $count\leftarrow 0$
\FOR{each $C_{i,t}$}
\FOR{each actant $t$ in $C_{i,t}$  and $K_j$}
\IF{$t$ in $K_j$}
\STATE $count\leftarrow count+1$
\STATE $Y_{gr}[t]\leftarrow j$
\STATE $Y_{pred}[t]\leftarrow i$
\ENDIF
\ENDFOR

\ENDFOR
\STATE $Pr_t\leftarrow count/length([t])$
\STATE $h_t\leftarrow Homogeneity(Y_{gr},Y_{pred})$ 
\STATE $c_t\leftarrow Completeness(Y_{gr},Y_{pred})$
\STATE $v_t\leftarrow V-Measure(Y_{gr},Y_{pred})$
\end{algorithmic}
\end{algorithm}

\section{Limitations}
There are certain limitations with our approach, including those related to data collection, the estimation of the narrative frameworks, the validation of the extracted narrative graphs, and the use of the pipeline to support real time analytics.

Data derived from social media sources tends to be very noisy, with considerable amounts of spam, extraneous and off topic conversations, as well as numerous links and images interspersed with meaningful textual data. Even with cleaning, a large number of text extractions are marred by spelling errors, grammatical errors, punctuation errors, and poor syntax. While these problems are largely addressed by our NLP modules, they produce less accurate entity and relationship extractions than those produced by our news corpora. Also, unlike news articles which tend to be well archived, social media posts, particularly on sites such as 4Chan, are unstable, often being deleted or hidden. Consequently, re-crawling a site can lead to the creation of substantively different target data sets. To address this particular challenge, we provide all of our data as an OSF repository \cite{shahsavari2020data}.

Visualizing large networks with clear labeling on nodes and edges is a well-known challenge in data visualization \cite{borner2019dagstuhl}. In the visualizations presented here, for purposes of clarity, we do not include the semantically rich labels on our edges. Consequently, the intra-community density based on context-dependent interactant relationships is difficult to discern, while the inter-community edges, which are traversed in discovering the narrative framework, can also be opaque. Although tabular representations of these relationships as presented here help address this problem, more localized network visualizations with clear edge labeling can be developed. Currently, edge labeled visualizations, such as those produced in Oligrapher, require a high degree of supervision to produce clear visualizations, and consequently producing them can be time intensive \cite{skom2019oligraph}.

The lack of consistent time stamping across and within social media sites makes determining the dynamics of the narrative frameworks undergirding social media posts difficult. In contrast to the news data harvested from the GDELT project, the social media data is marked by a very coarse time frame due to inconsistent time stamps or no time stamps whatsoever. Comparing a crawl from one day to the next to determine change in the social media forums may help attenuate this problem. Given the potential for significant changes due to the deletion of earlier posts, or the move of entire conversations to different platforms, the effectiveness of this type of strategy is greatly reduced. Because of the limited availability of consistently time-stamped data, our current comparison between the social media conspiracy theory narrative frameworks, and those appearing in the news, is limited to a three week window.

There appears to be a fairly active interaction between the ``Twittersphere'' and other parts of the social media landscape, particularly Facebook. Many Tweets, for instance, point to discussions on social media and, in particular, on Facebook. Yet, because of the restrictions on access to Facebook data for research purposes, we are unable to consider this phenomenon. Future work will incorporate Tweets that link to rumors and other conspiracy theories in our target social media arena. As part of this integration, we also plan to include considerations of the trustworthiness of various twitter nodes, and the amplification role that ``bots'' can play in the spread of these stories \cite{ferrara2020twitbots} \cite{davis2016bots}.

As with a great deal of work on social media, there is no clear ground truth against which to evaluate or validate. This problem is particularly apparent in the context of folkloric genres such as rumor, legend and conspiracy theories, as there is no canonical version of any particular story. Indeed, since folklore is always a dynamically negotiated process, and predicated on the concept of variation, it is not clear what the ground truth of any of these narratives might be. To address this problem, we consider the narrative frameworks emerging from social media and compare them to those arising in the news media. The validation of our results confirms that our social media graphs are accurate when compared to those derived from news media.

Currently, our pipeline only works with English language materials. The modular nature of the pipeline, however, allows for the inclusion of language-specific NLP tools, for parsing of languages such as Italian or Korean, both areas hard hit by the pandemic, and likely to harbor their own rumors and conspiracy theories.

\section{Results and Evaluation}

\begin{figure}
    \centering
    \includegraphics[width=\textwidth]{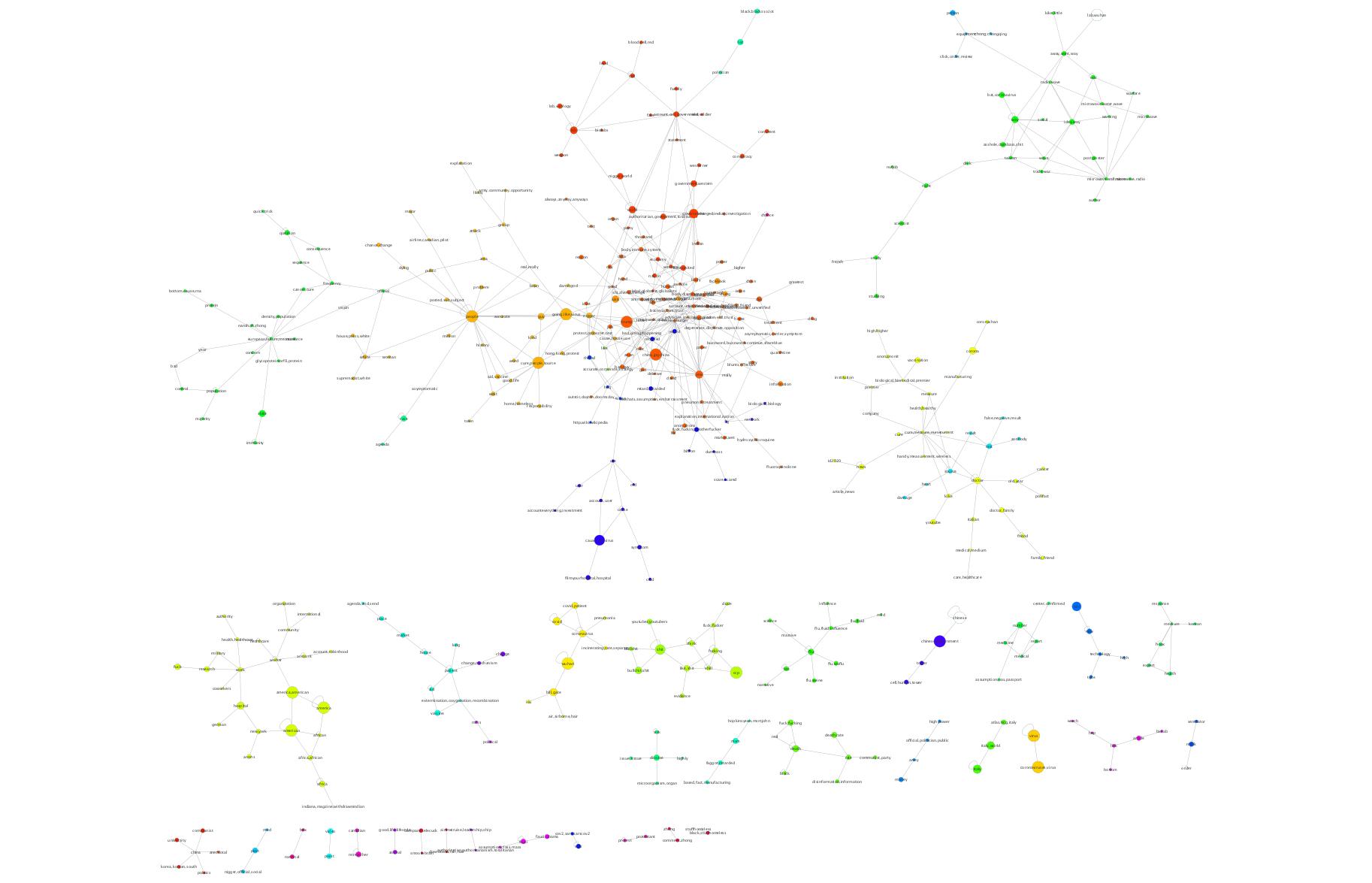}
    \caption{\textbf {Overview graph of communities in social media corpus.} Nodes are colored by community, and sized by NER score. Even small communities may include subnodes with very high NER scores. Narrative frameworks are drawn from these communities, each of which describes a knowledge domain in the conversation. Nodes with multiple community assignments are colored white, and appear with their highest ranked community. An overarching narrative framework for a conspiracy theory often aligns subnodes from numerous domains.}
    \label{fig:socmed_communities}
\end{figure}

After running the pipeline and community detection, we find a total of fifty-two  communities constituting the various knowledge domains in the social media corpus from which actants and interactant relationships are drawn to create narrative frameworks (See Figure \ref{fig:socmed_communities}). Table \ref{tab:communities} shows the largest of these communities with a subnode count $\geq 9$; the temporary labels for each community are based on an aggregation of the labels of the nodes with the highest NER scores in those communities. 

\begin{table}[]
\centering
\caption{The largest eleven communities in the social media corpus in descending order of size. The labels are derived from the subnode labels for the three nodes with highest degree in each community.}
\label{tab:communities}
\begin{tabular}{@{}lll@{}}
\toprule
Community number & Community Size & Community label \\ \midrule
2 & 66 & China, Trump, Jew \\
5 & 39 & People, cure, virus \\
8 & 28 & corona, doctor, vaccination \\
1 & 27 & government, lab, world \\
16 & 26 & bat, coronavirus, idea \\
9 & 25 & America(n), hospital(s), Africa \\
17 & 21 & state, question, population \\
15 & 11 & flu, bat, science \\
35 & 10 & cause, film your hospital, symptom(s) \\
7 & 9 & Wuhan, Covid, Bill Gates \\
24 & 9 & vaccine, patient, market \\ \bottomrule
\end{tabular}
\end{table}

A narrative framework for a conspiracy theory, which may initially take shape as a series of loosely connected statements, rumors and innuendo, is composed from a selection of subnodes from one or more of these communities and their intra- and inter-community relationships. It is worth noting that several of the communities are ``meta-narrative'' communities, and describe either the social media platform or linguistic peculiarities of communications on that platform (e.g. community 34), the background for the discussion itself (e.g. community 10 which focuses on Italy, one of the European hot-spots for the virus, or community 31 which focuses on medical equipment such as masks and ventilators), or other platform specific discussions (e.g. communities 4 and 11 which focus on Facebook and YouTube respectively). These background and ``meta-narrative'' communities, after thresholding, tend to be quite small, with an average of $3.6$ subnodes per community. Nevertheless, several of them include subnodes with very high NER scores, such as community 6, with only two nodes, ``Corona virus'' and ``virus'', meaning this community is very likely to be included as part of more elaborated conspiracy theory narrative frameworks. The five largest communities, in contrast, range in size from 26 to 66 nodes. These five communities, along with several other large communities, form the main reservoir of actants and interactant relationships for the creation of conspiracy theory narrative frameworks.

\begin{figure}
    \centering
    \includegraphics[scale=0.2]{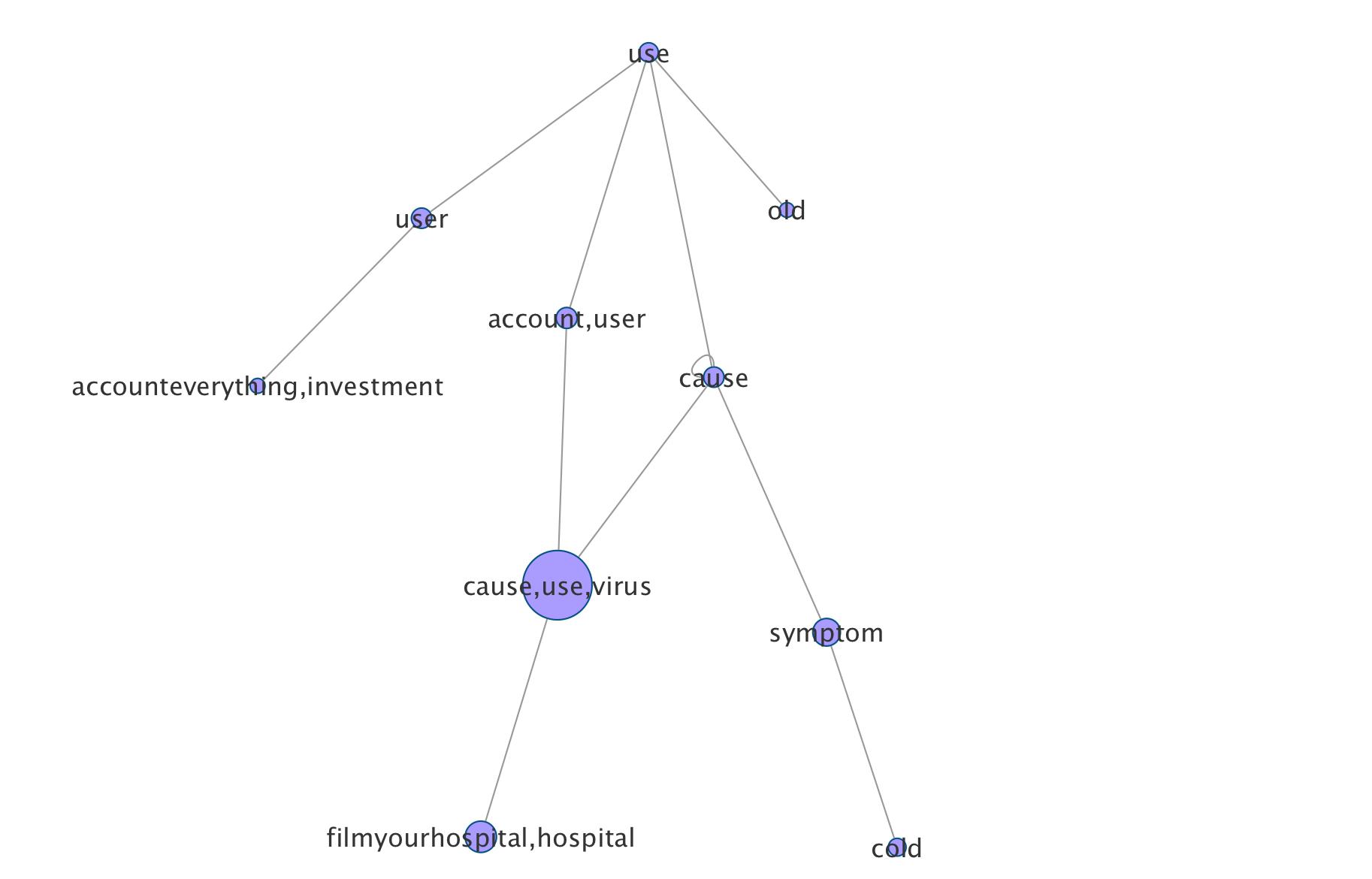}
    \caption{\textbf {Graph of a community with the ``Film your hospital'' movement actant subnode, and the implication that the virus only causes mild, cold-like symptoms.}} 
    \label{fig:film_hospital}
\end{figure}

Each community represents a group of closely connected actant subnodes with those connections based on the context-dependent interactant relationships. Traversing paths based on these inter-actant relationships within and across communities highlights how members posting to the forums understand the overall discussion space, and provide insight into the negotiation process concerning main actants and interactant relationships. For example, looking solely within a moderately sized community in the graph, such as community 32 with ten subnodes, one can easily see the connection that the members of these forums have made between the cold symptoms caused by the virus and the "film your hospital" movement (See Figure \ref{fig:film_hospital}). This movement encourages people to film their local hospitals as ``proof" that the hospitals are neither overcrowded nor being inundated with acutely ill virus sufferers, and has been compared to the ``truther'' movement that challenged the legitimacy of Barack Obama's birth certificate \cite{sommer2020truthers}. Consequently, subnodes from this community are activated in the context of narrative frameworks based on these ``truther'' type conspiracy theories, and the narrative frameworks that propose that the pandemic is largely a hoax.

\begin{figure}
    \centering
    \includegraphics[scale=0.27]{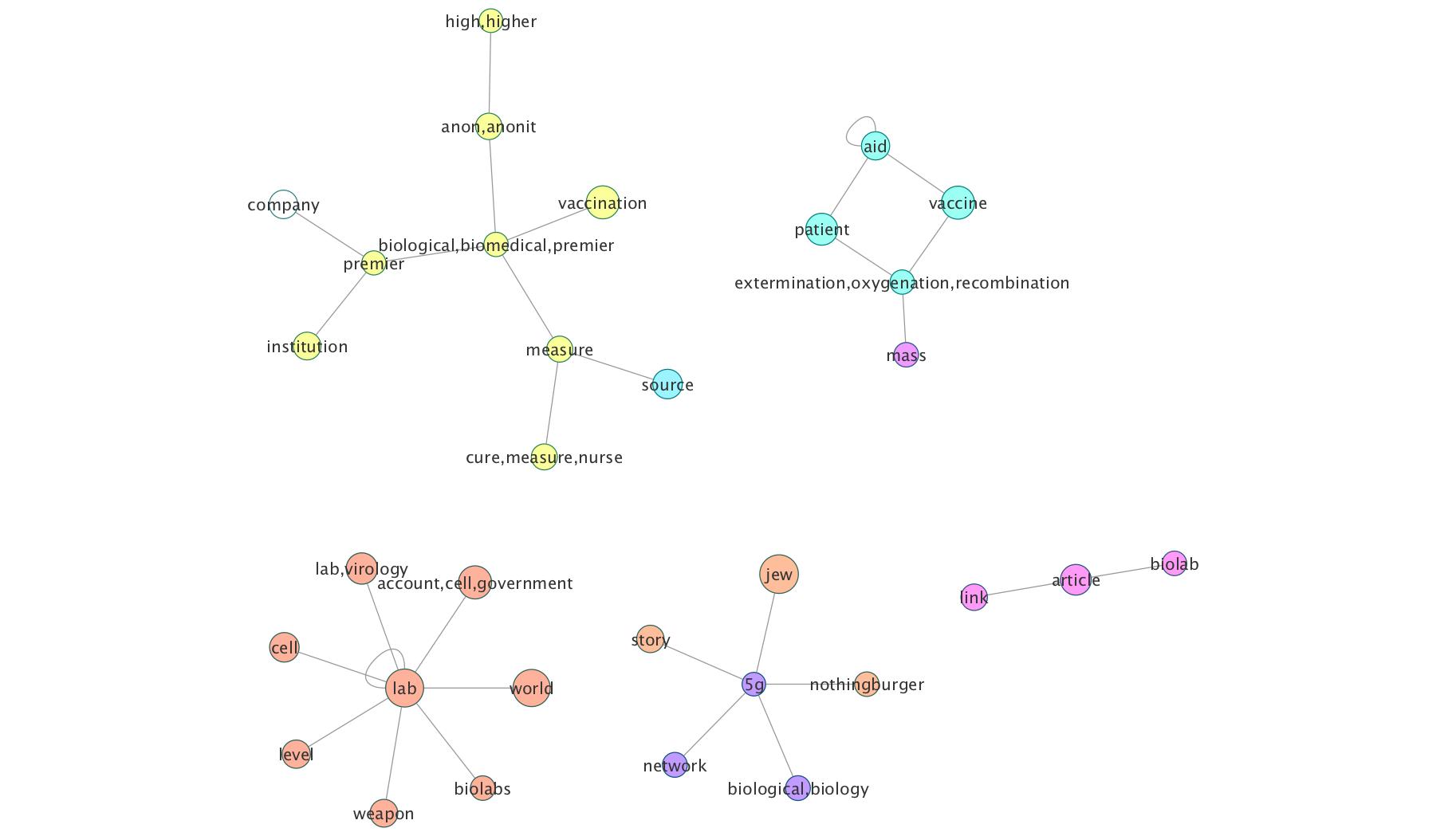}
    \caption{\textbf {Graph of a series of communities based on seed words with vaccines or vaccination, and biomedicine.} The appearance of the 5G node may suggest that the forum members are aligning disparate narrative frameworks into a single, totalizing narrative.} 
    \label{fig:vaccine_exterm}
\end{figure}

Traversing communities based on the relationship paths between communities, or based on subnodes with similar high ranked labels, provides opportunities to find other emerging narrative frameworks. Starting with vaccines and vaccination from communities 8 and 24, and bio-medicine related labels from communities 1, 8, 37 and 43, it becomes apparent that there is a narrative framework supporting stories that posit vaccination as a biological weapon used to exterminate segments of the population \ref{fig:vaccine_exterm}. Intriguingly, this small selection of nodes also brings in the 5G node, which is an important node for a separate narrative framework that joins together cellular networks and microwaves as key technologies for activating the virus as a biological weapon. It also brings in the Bill Gates node; Gates is a key actant in a narrative framework suggesting that his support for widespread vaccination programs is a cover for his alleged desire for a global surveillance apparatus (See Table \ref{tab:bill_news}).

\begin{figure}
	\centering 
	\includegraphics[width=0.8\textwidth]{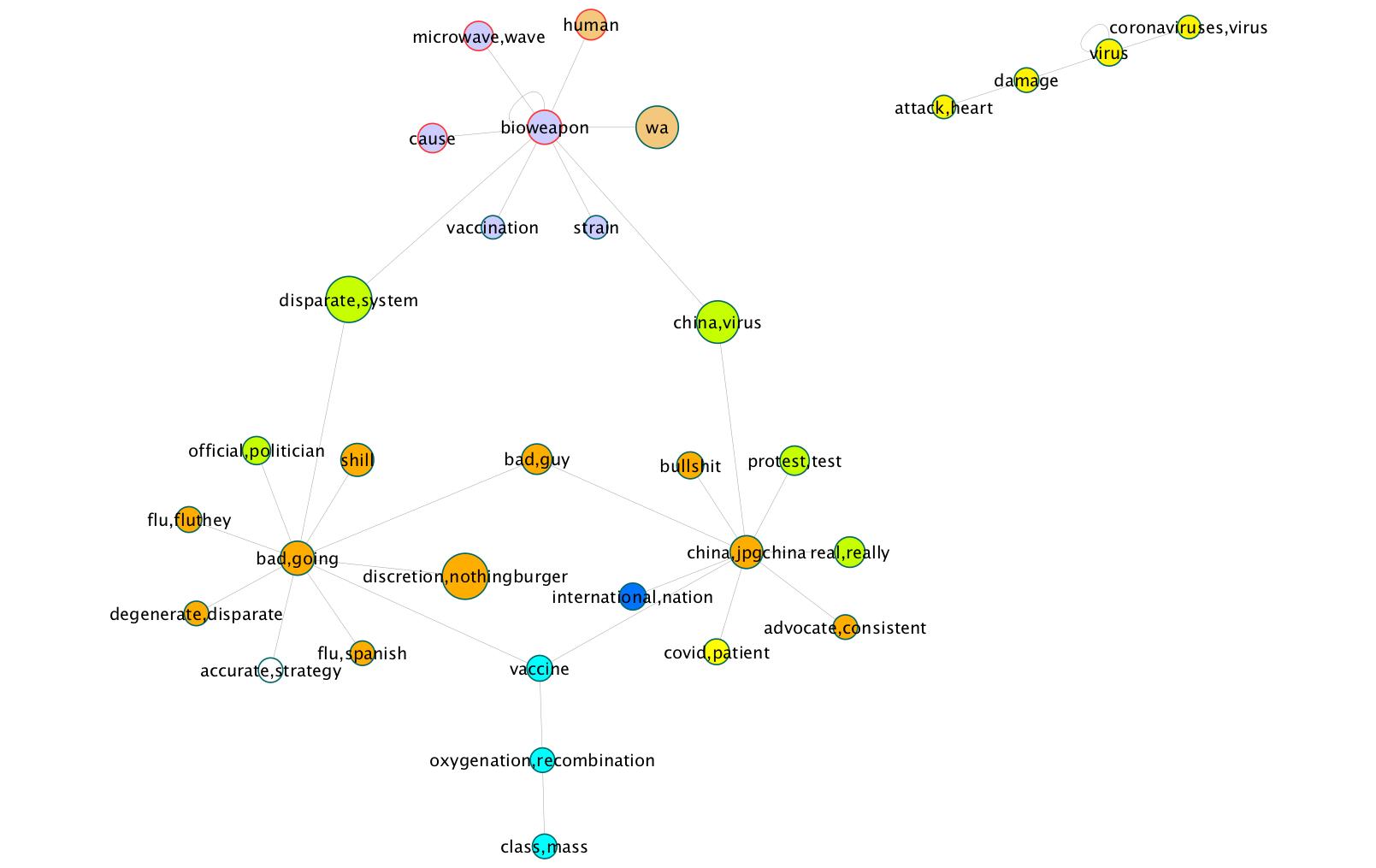}
	\caption{ {\textbf{A subgraph showing multiple pathways through actants}: Traversing the graph by selecting high degree nodes in each community and their nearest neighbors in the broader graph provide candidate narrative frameworks.}
		\label{fig:china_fork}}
\end{figure}

To assemble candidate narrative frameworks, we consider the subnodes with both the highest degree and/or the highest NER score in any community, and select those that are not explicitly generic (e.g. ``people''). We then find the nearest neighbors of those high degree subnodes, and traverse the graph to the next nearest neighbors. Each edge in this subgraph can be labeled, providing alternate pathways across the subgraph. For example, a subgraph drawing on subnodes from six different communities proposes several possibilities: (i) that the Covid-19 pandemic is a hoax, and not much more dangerous than the flu, a product of the bio-weapons research of China and an outcome of their system of government, and will be addressed by a vaccine like that for the flu, or (ii) the pandemic is a Chinese virus related to bio-weapons, and may be resistant to vaccination; in either case, there are bad actors behind it (See Figure \ref{fig:china_fork}).  This subgraph then constitutes two candidate narrative frameworks. Iterating this process across the graph, we are able to identify numerous narrative frameworks in the social media data, which may be combined into larger, connected narrative frameworks.

\begin{figure}
	\centering 
	\includegraphics[width=0.90\linewidth,height=0.90\textheight]{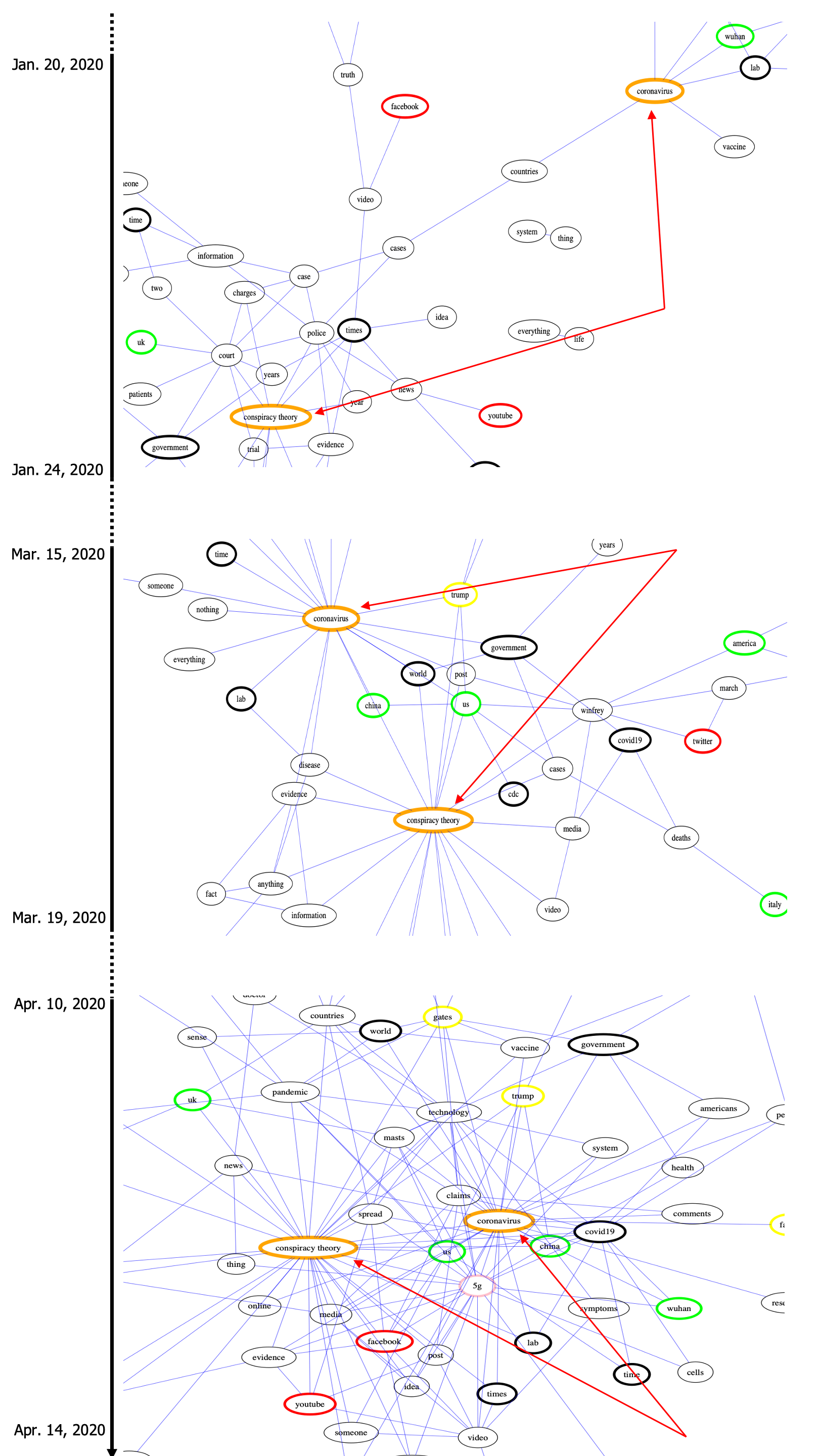}
	\caption{ {\textbf{Progressive attachment of ``coronavirus'' to ``conspiracy theory'' in the co-occurrence network of news reports conditioned on entities found in social media}: The orange-outlined nodes represent the two concepts, as they gravitate toward one another over time and form new simple paths - from left to right; 5-day intervals starting on January 20, 2020, March 15, 2020, and April 10, 2020. Celebrities are in yellow, media outlets in red, current conspiracy theories' keywords in pink (manually colored), places in green and corporations/entities in black}
		\label{fig:news_timeline_of_2nodesa}}
\end{figure}

The relationship between the discussions occurring in social media, and the reporting on conspiracy theories in the media, changed over our study period. In mid to late January, when the Corona virus outbreak appeared to be limited to the central Chinese city of Wuhan, and of little threat to the United States, news media reporting on conspiracy theories
had very little connection to reporting on the Corona virus outbreak.
As the outbreak continued through March 2020, the reporting on conspiracy theories gradually moved closer to the reporting on the broader outbreak. By the middle of April, reporting on the conspiracy theories being discussed in social media, such as in our research corpus, had moved to a central position (see Figure \ref{fig:news_timeline_of_2nodesa}). 

\begin{figure}
	\centering 
	\includegraphics[width=0.75\textwidth]{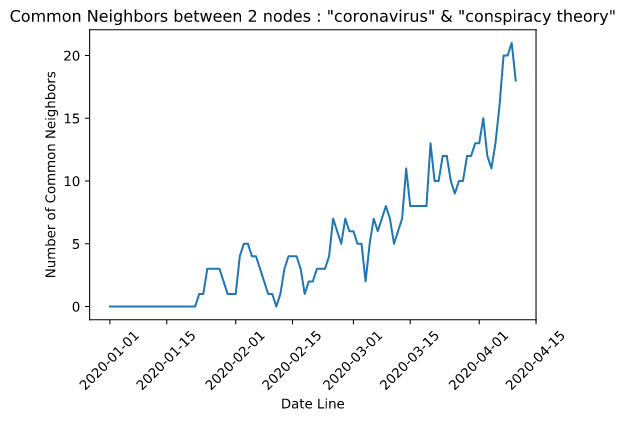}
	\caption{ {\textbf{Number of common neighbors between ``coronavirus'' and ``conspiracy theory'' over time in the news reports}: Across all 101 segments of 5-day intervals, the number of simple paths empirically increases exponentially suggesting the closer ties between the two entities across time}
		\label{fig:news_timeline_of_2nodesb}}
\end{figure}

The connection between these two central concepts in the news---``coronavirus'' and ``conspiracy theory'' can also be seen in the rapid increase in the shared neighbors of these subnodes (defined in Equation \ref{eqn:NCN}) in the overall news graph during the period of study (see Figure \ref{fig:news_timeline_of_2nodesb}).

\begin{figure}
	\centering
  \begin{subfigure}{1.0\textwidth}
	\includegraphics[width=\textwidth]{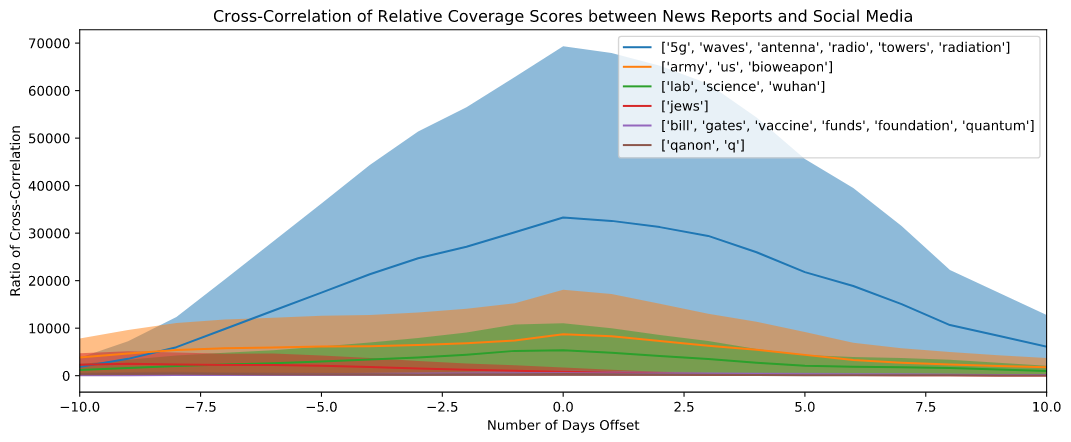}
	\caption{Words in a community are matched to words present in the news reports and social media. Both the news reports and social media are smoothed for 5-day intervals. The mean and standard deviation are computed per time stamp and marked.}
	\end{subfigure}\par\bigskip
   \begin{subfigure}{1.0\linewidth}
	\includegraphics[width=\textwidth]{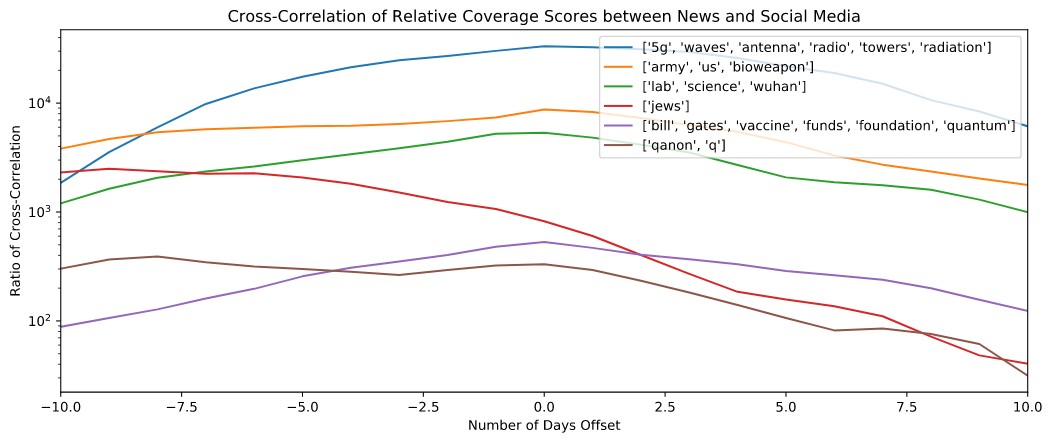}
	\caption{Mean trajectories show the relative differentiation of each community. Note the scaling on the y-axis is logarithmic.}
	\end{subfigure}
	\caption{\textbf{Ratio of Cross-Correlation of Relative Coverage Score for Word-Level Community Hits in social media against the news reports}: The mean and standard deviation of the relative coverage score are computed per time stamp across 20 trials with 500 community members each}
	\label{fig:news_timeline_of_nodesc}
\end{figure}

Since our dataset contains dated 4Chan \textit{and} GDELT data from March 28, 2020 to April 14, 2020, communities from the social media corpus were explored within the subset of news media between the same dates using Coverage Scores defined in Equation \ref{eqn:CS}. The cross-correlation of the ratio of coverage scores for different fixed communities, to a random community is provided in Figure \ref{fig:news_timeline_of_nodesc}. 

The higher average scores for the ``5G'' community including words such as \{"5g", "waves", "antenna", "radio", "towers", "radiation"\}, suggests that this community was matched more frequently than other communities compared to a baseline random community. A peak at zero days offset within the time period from March 28, 2020 to April 14, 2020 implies that the news reports are correlated in time to 4Chan thread activity. In addition, these plots suggest that few communities dominate conspiracy theories more than others. The viability of other communities such as \{"army", "us", "bioweapon"\} and \{"lab", "science", "wuhan"\} suggest the lack of a single dominant conspiracy theory \textit{consensus} narrative. Instead, it appears that numerous conspiracy theories may be vying for attention.

\begin{figure}[ht] 
  \begin{subfigure}[b]{0.5\linewidth}
    \centering
    \includegraphics[width=1.0\linewidth]{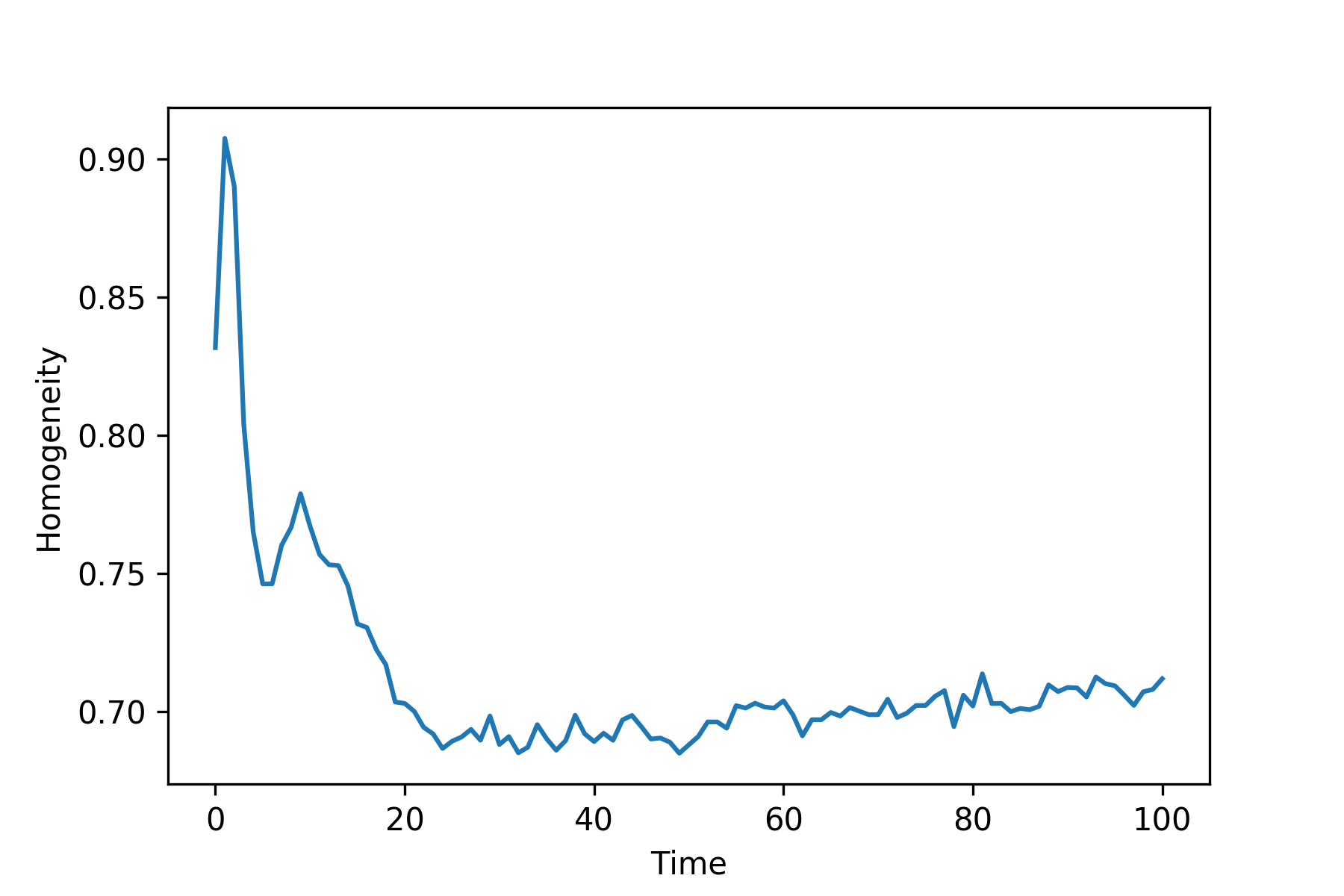} 
    \caption{ {\textbf{Homogeneity of News based communities}} }
  \end{subfigure}
  \hfill
  \begin{subfigure}[b]{0.5\linewidth}
    \centering
    \includegraphics[width=1.0\linewidth]{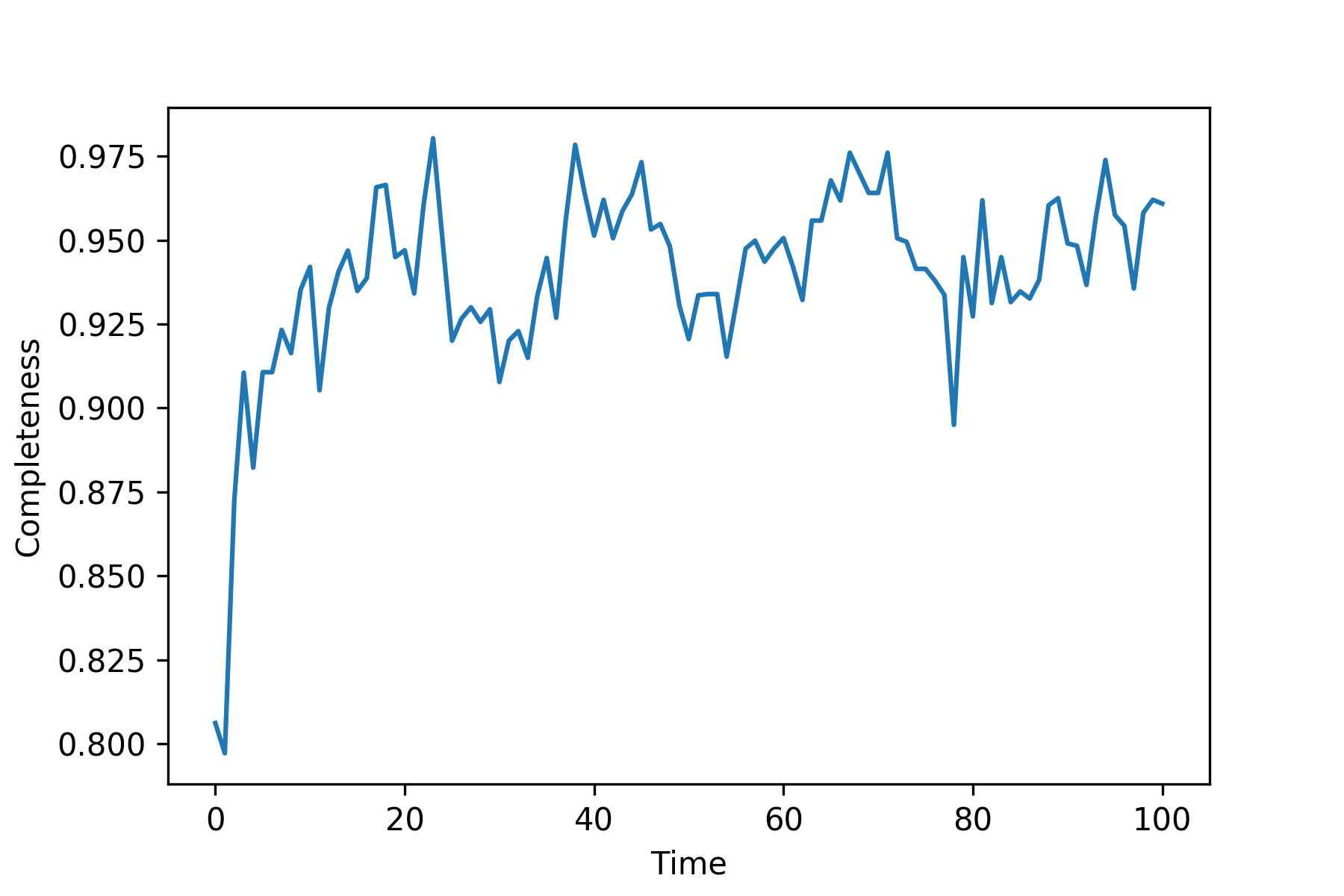} 
    \caption{ {\textbf{Completeness of News based communities}} }
  \end{subfigure} 
  \vskip\baselineskip
  \begin{subfigure}[b]{0.5\linewidth}
    \centering
    \includegraphics[width=1.0\linewidth]{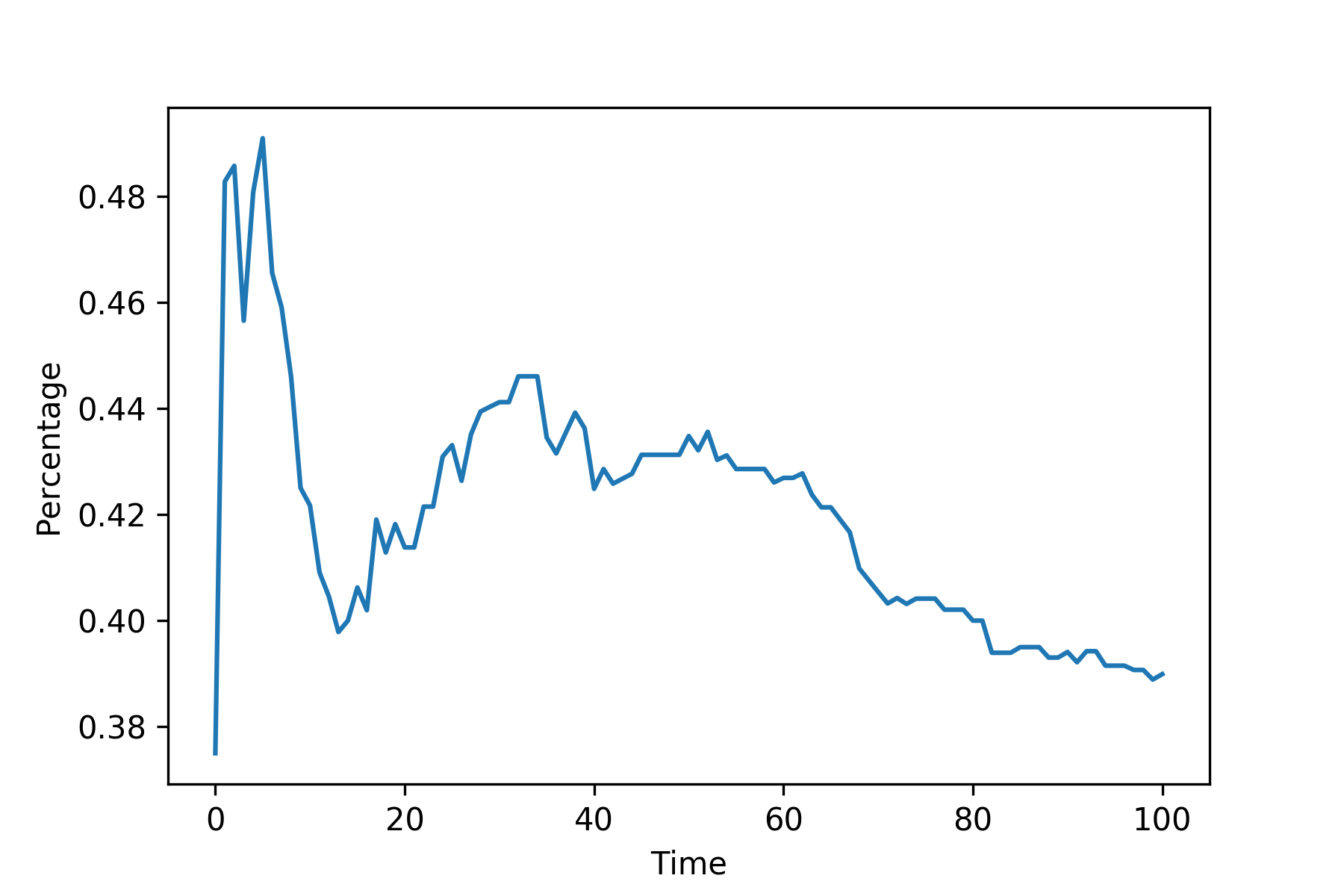} 
    \caption{ {\textbf{Percentage of coverage}}}
  \end{subfigure}
  \quad
  \begin{subfigure}[b]{0.5\linewidth}
    \centering
    \includegraphics[width=1.0\linewidth]{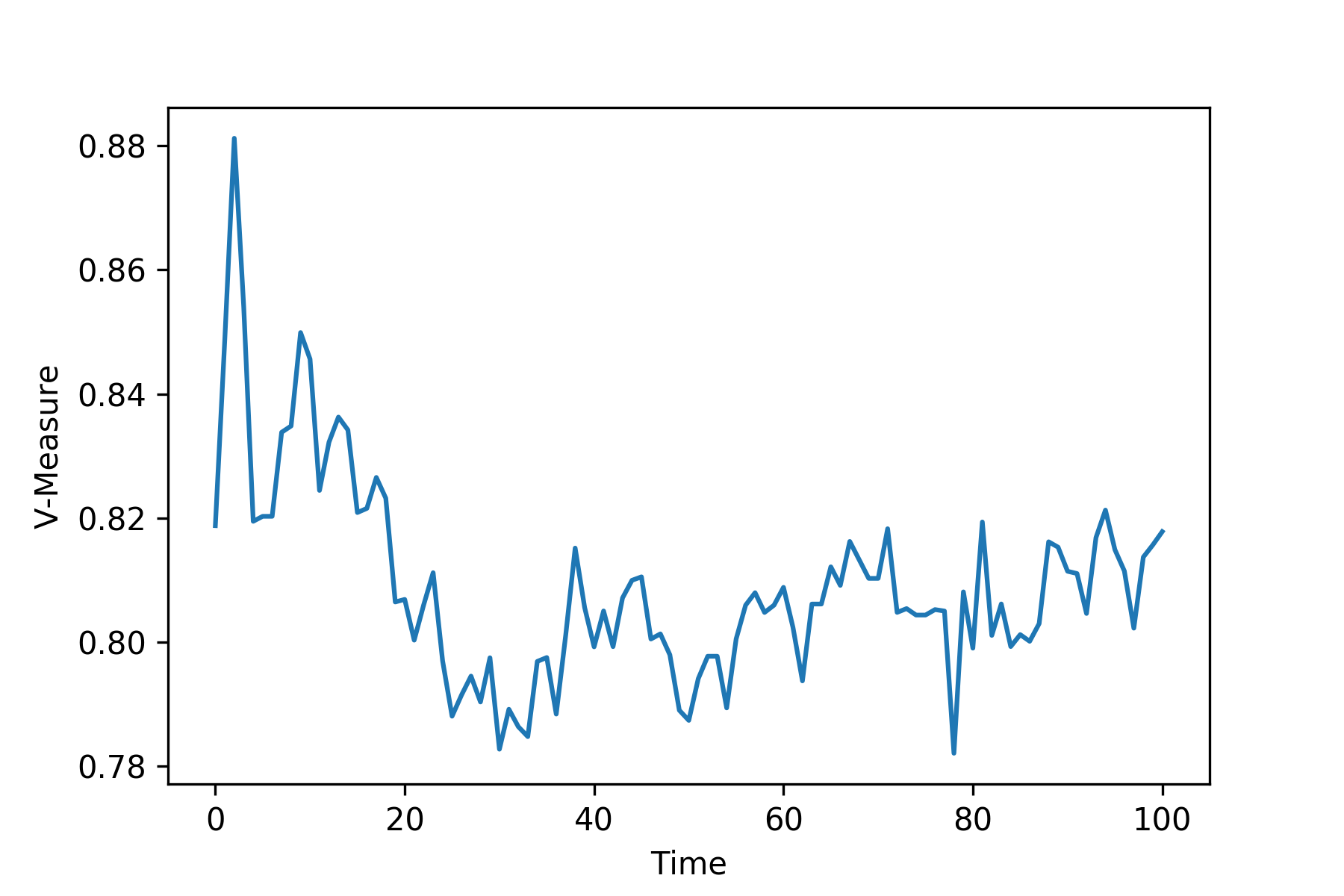} 
    \caption{\textbf{V-Measure}} 
   
  \end{subfigure}
  \caption{Scores of (a) Homogeneity, (b) Completeness, (c) Coverage and (d) V-Measure are provided to compare News based communities with Social Media communities.
  Here we used $Y_{pred}$ and $Y_{gr}$ derived in algorithm \ref{alg:algo_comm} as our cluster label and classes. Completeness measures how members of a given class are assigned to the same cluster, while homogeneity measures how each cluster contains only members of a single class. Their harmonic mean is the V-Measure \cite{sklearn_link}.
  }
  \label{comm_eval}
\end{figure}





We examine ``bill gates'', a key actor frequently found in the common neighbors set between ``coronavirus'' and ``conspiracy theory''. Key relationships extracted by our pipeline on the news reports provide a qualitative overview of the emergence of ``bill gates'' as a key actor (see Table \ref{tab:bill_news}).

\par
Last, the evaluations based on algorithm \ref{alg:algo_comm} are shown in Figure \ref{comm_eval}. The plots indicate the saturation of completeness and homogeneity saturates at $\sim 72\%$ across time. Similarly, the V-measure saturates at $\sim 80\%$. These scores per time sample, represent the fidelity of the process of cluster matching.

\begin{table}[]
\centering
\caption{\textbf{A qualitative overview of key relationships that refer to ``Bill Gates'' in social media and the news reports.} These relationships describe the influence of Bill Gates in connecting the Corona virus to conspiracy theories.}
\begin{tabular}{|l|l|l|}
\hline
\textbf{Date} & \textbf{News Reports} & \textbf{4Chan Threads} \\ \hline

04/04 & \begin{tabular}[c]{@{}l@{}}{[}Bill \{Gates\}{]} $\rightarrow$ {[}\{predicted\}{]}\\  $\rightarrow$ {[}the \{outbreak\} and the China biolabs{]}\end{tabular}                                                         & \begin{tabular}[c]{@{}l@{}}{[}\{5g\}{]} $\rightarrow$ {[}\{causes\}{]} $\rightarrow$ {[}\{coronavirus\}{]}, \\ {[}regular \{people\}{]} $\rightarrow$ [\{go\}] $\rightarrow$ {[}\{untested\}{]}\end{tabular} \\ \hline
04/07 & \begin{tabular}[c]{@{}l@{}}{[}conspiracy theorist David \{Icke\}{]} $\rightarrow$ \\ {[}\{added\}{]} $\rightarrow${[}that Bill \{Gates\}, who\\  is helping fund vaccine research, should be jailed{]}\end{tabular} & \begin{tabular}[c]{@{}l@{}}{[}\{Gates\}{]} $\rightarrow$ {[}\{saying\}{]} $\rightarrow$ \\ {[}{[}...{]}we all {[}...{]} accept his discount mark of the beast{]}\end{tabular}                             \\ \hline
04/09 & \begin{tabular}[c]{@{}l@{}}{[}Bill \{Gates\}{]} $\rightarrow${[}\{invented\}{]} \\ $\rightarrow${[}\{5G\} to depopulate the world{]}\end{tabular}                                                                  & \begin{tabular}[c]{@{}l@{}}{[}the satanic \{cabal\}{]} $\rightarrow$ {[}to \{leverage\}{]} $\rightarrow$ \\ {[}crisis into a forced vaccination /I D \{program\}{]}\end{tabular}                          \\ \hline
\end{tabular}
\label{tab:bill_news}
\end{table}

\section{Discussion}

We discover a series of important phenomena concerning the (i) narrative frameworks that undergird conspiracy theories and their constituent rumors circulating on and across social media, and (ii) the interaction between social media and the news. The lack of authoritative information about the Covid-19 pandemic has allowed people to provide numerous, varied explanations for its provenance, its pathology, and both medical and social responses to it. These conversations are not occurring in isolation, but rather cross various social media platforms and also interact with news reporting on the pandemic as it unfolds. Similarly, journalists are keenly aware of the discussions occurring in social media.

The main connected component for the aggregated social media corpus reveals the centrality of several significant conspiracy theories: (i) that the 5G cellular network is a root cause of the virus and with the lethality of the virus tied to its role as a bio-weapon; (ii) that the virus is linked to laboratories in China. These overlapping narratives are set against a broader discussion of the viruses rapid emergence in Washington state, and the role that Trump has played in responding to the virus.

\begin{figure}
	\centering 
	\includegraphics[width=\textwidth]{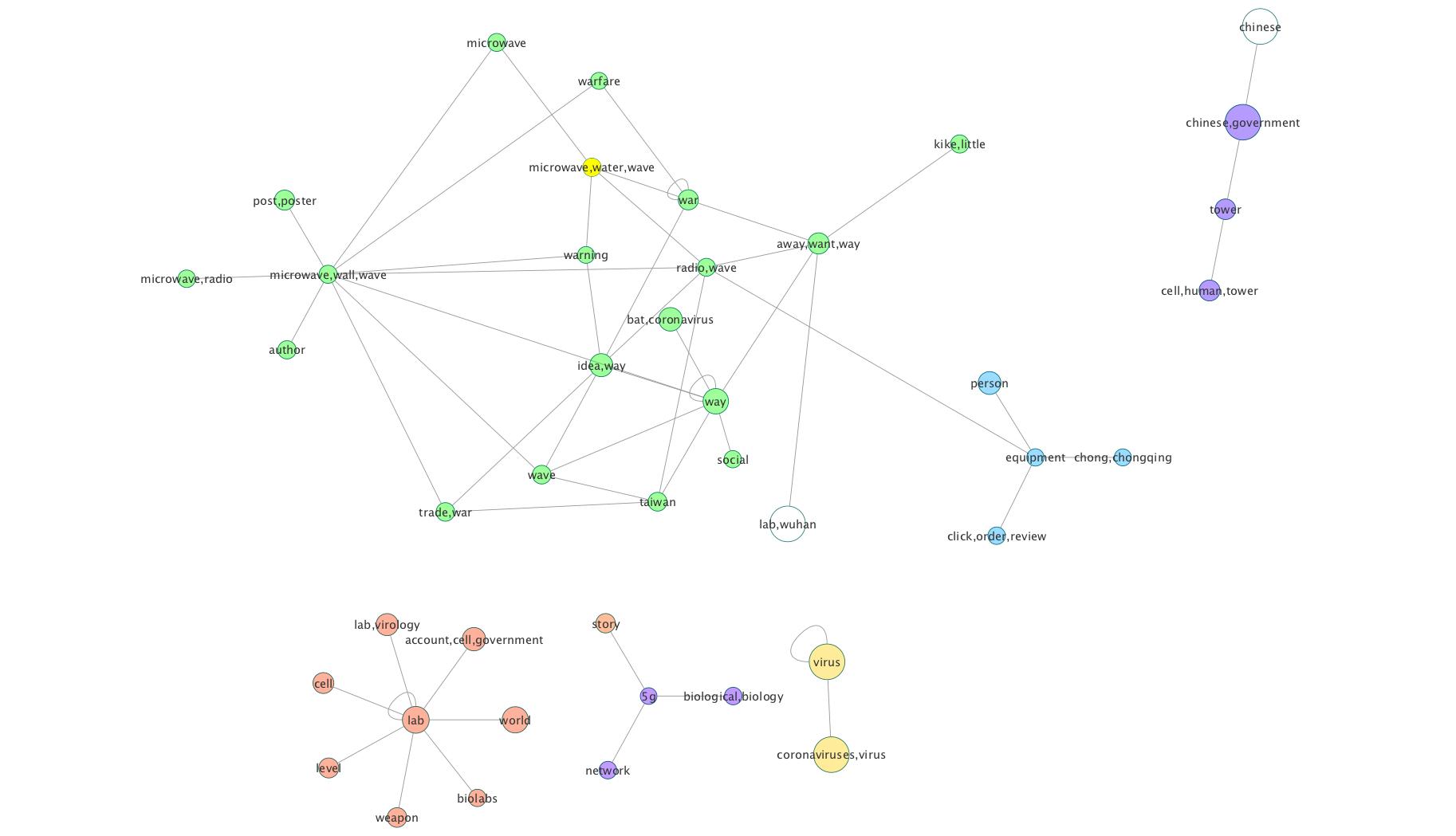}
	\caption{ {\textbf{Communities comprising the 5g narrative framework:} The narrative framework of microwave radiation emanating from the 5g network activates the bioweapon of the corona virus. Node color is assigned by community. Nodes have been scaled by NER ranking.}
		\label{fig:5g}}
\end{figure}

Traversing the communities across nearest neighbors makes apparent the various conspiracy narrative frameworks active in the social media data. The 5G cellular conspiracy, for instance, draws from five communities that, taken together, provide a clear understanding of the underlying framework of this conspiracy theory (See Figure \ref{fig:5g}). In this conspiracy theory, radiation from 5G cellular networks, using Chinese equipment and related to the broader US trade war with China, allows the virus, developed possibly as a bio-weapon in a Wuhan laboratory, to spread unhampered with devastating effect through human communities.

\begin{figure}
	\centering 
	\includegraphics[width=\textwidth]{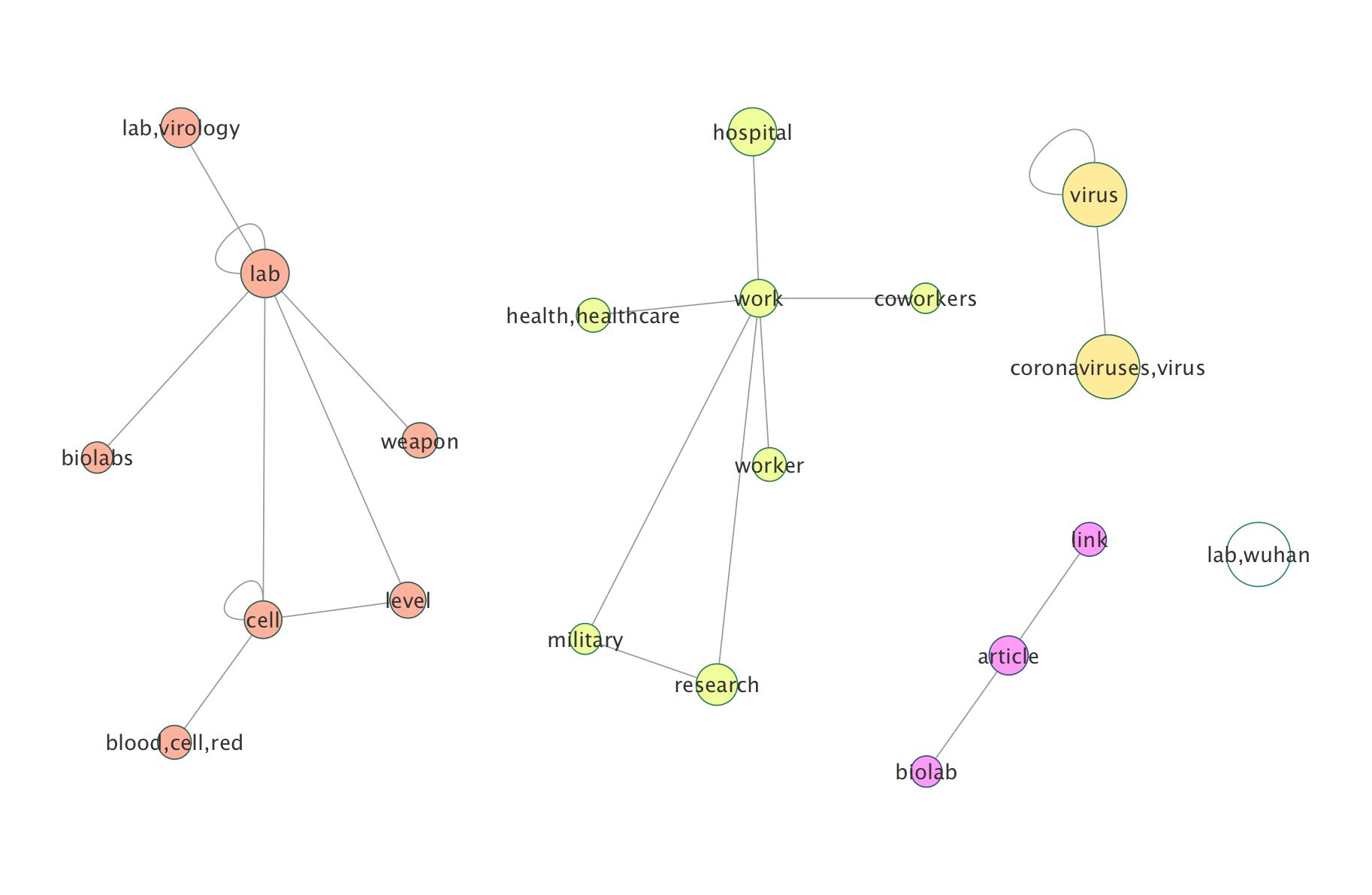}
	\caption{ {\textbf{Communities comprising the Covid-19-as-bioweapon narrative framework.} The narrative framework focuses heavily on laboratories and the potential role of the virus as a weapon.}
		\label{fig:bio-weapon}}
\end{figure}

Another conspiracy theory that competes for attention in the overall social media space is one suggesting that the Covid-19 virus escaped from a Chinese biological weapons laboratory, most likely in Wuhan, as part of either a deliberate release or accidentally, and is closely related to the 5G conspiracy theory detailed above (See Figure \ref{fig:bio-weapon}).

\begin{figure}
	\centering 
	\includegraphics[width=\textwidth]{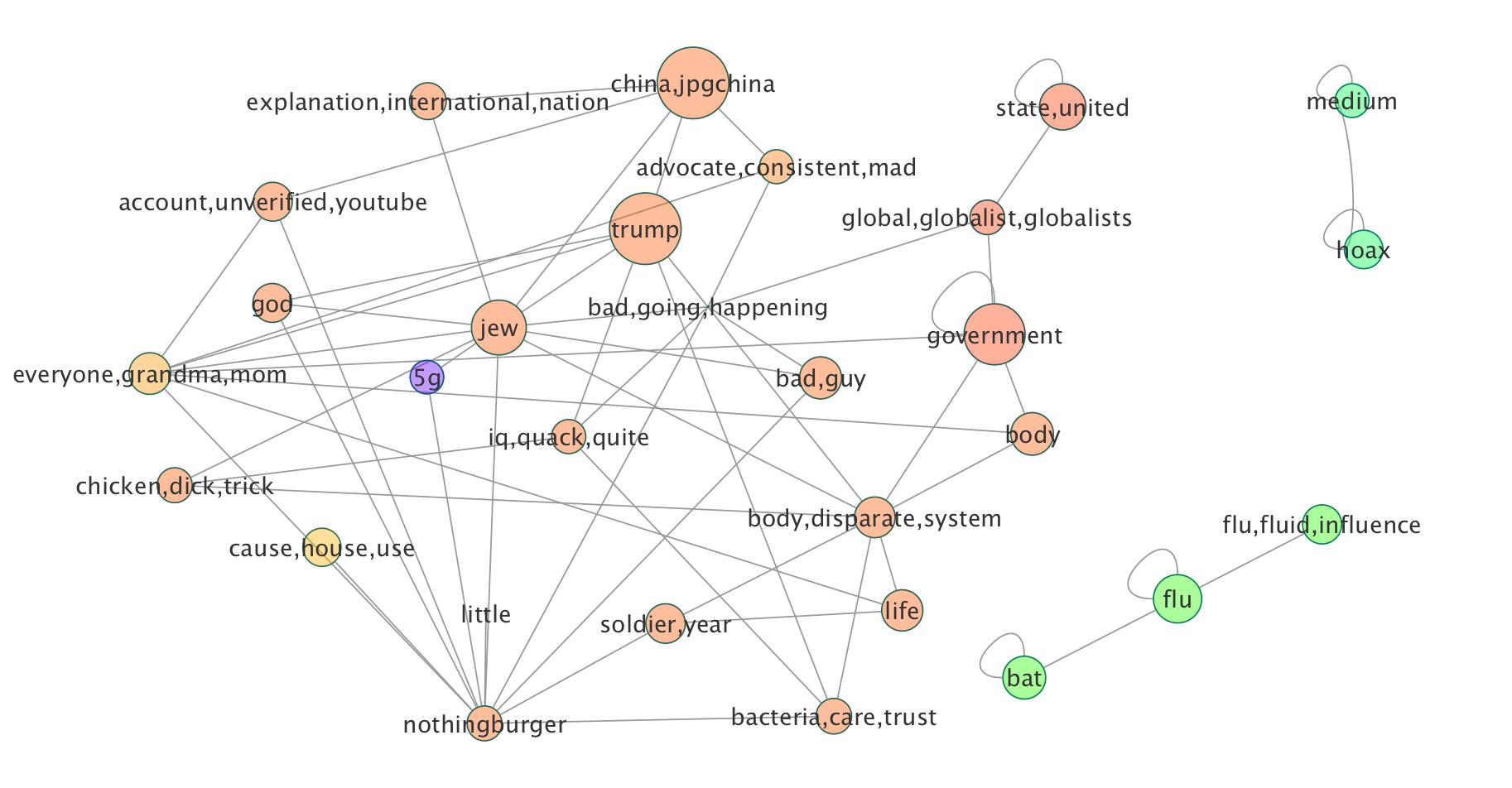}
	\caption{ {\textbf{The communities comprising the globalist hoax narrative framework:} Here, a globalist cabal has conspired to foist the hoax of the Corona virus on the world, with the virus presenting with mild flu-like symptoms.}
		\label{fig:globalist_hoax}}
\end{figure}

Another narrative framework proposes that the pandemic is a hoax. This framework includes actants such as Trump and the Republicans who are fighting against the globalists, Democrats and, in keeping with the long history of anti-Semitic conspiracies, the ``Jews", all of whom have conspired to perpetrate this hoax. While the goal of the hoax is not made explicit, it is clear that the virus presents with nothing more than mild symptoms, and is no more dangerous than the flu (See Figure \ref{fig:globalist_hoax}). That the pandemic is a hoax inspires the ``film your hospital'' movement as a means for publicizing the ``discovery'' that the virus poses no meaningful threat, other than the economic threat of stay-at-home orders.

\begin{figure}
	\centering 
	\includegraphics[width=\textwidth]{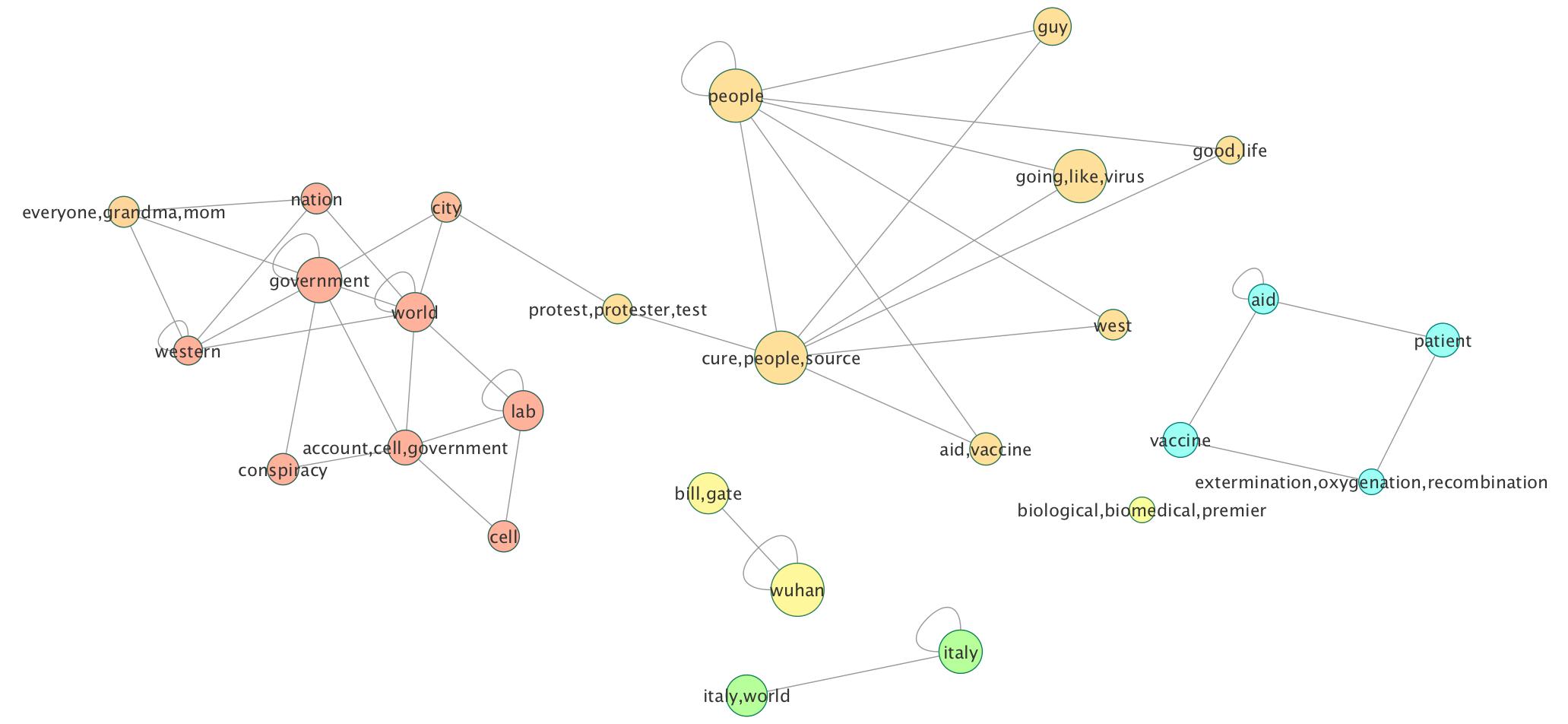}
	\caption{ {\textbf{The communities comprising the Bill Gates - Vaccination - global surveillance narrative framework:} The Gates foundation, under the guise of a Covid-19 vaccination program, intends to develop a method for global surveillance.}
		\label{fig:gates}}
\end{figure}

A final narrative framework includes Bill Gates, and his foundation, as central actors in a conspiracy involving global vaccination and surveillance. In the discussion forums, posters discuss Gates's desire to apply ``quantum tattoos'' as part of a broader effort at global surveillance under the guise of a vaccination campaign to combat the Corona virus (See Figure \ref{fig:gates}).

There are other nucleations of conspiracy theory narrative frameworks in the overall social media space. Some of these are likely to be subsumed by the main competing conspiracy theory narrative frameworks as the process of framework alignment---a key aspect of the totalizing nature of conspiracy theories, which attempt to fit all events into a single belief framework---comes into play. This monological thinking appears to be at work in the alignment of the 5G conspiracy theory with the biological weapons conspiracy theory, with both of those frameworks sharing close connections with the narrative framework describing the pandemic as a whole. They also emphasize the role that Trump, the military, and the state of Washington all play in the outbreak.

In earlier work on conspiracy theories, we discovered that conspiracy theorists collaboratively negotiate a single explanatory narrative framework, often composed of a pastiche of smaller narratives, aligning otherwise unaligned domains of human interaction as they develop a totalizing narrative \cite{samory2018conspiracy}\cite{goertzel1994belief}\cite{tangherlini_plos}. In many conspiracy theories, this coalescence of disparate stories into a single explanatory conspiracy theory relies on the conspiracy theorists' self-reported access to hidden, secret, or otherwise inaccessible information. They then use this information to generate ``authoritative'' links between disparate domains, engaging in what Goertzel has labeled ``monological thinking'' \cite{goertzel1994belief}. 

For the current pandemic, however, a single unifying corpus of special or secret knowledge does not yet exist---there are no ``smoking guns'' to which the conspiracy theorists can point. Consequently the social media space is crowded by a series of potentially explanatory conspiracy theories. In the various forums we considered, proponents of different narratives fight for attention, while also trying to align the disparate sets of actants and interactant relationships in a manner that allows for a single narrative framework to dominate and, by extension, to provide the ``winning" theorists with the bragging rights of having uncovered ``what is really going on."

\section{Conclusion}

As the global Covid-19 pandemic continues to severely challenge societies across the globe, and as access to accurate information both about the virus itself and what lies in store for our communities continues to be limited, the generation of rumors and conspiracy theories with some explanatory value will continue unabated. Although news media have paid considerable attention to the well-known Q-Anon conspiracy theory (perhaps the most capacious of conspiracy theories of the Trump presidency), the social media conversations have focused on four main conspiracy theories: (i) the virus as related to the 5G network, explaining both the Chinese provenance of the virus through the connection to the communications giant Huawei; (ii) the release, either accidental or deliberate of the virus from, alternately, a Chinese laboratory or an unspecified military laboratory, and its role as a bio-weapon;  (iii) the perpetration of a hoax by a globalist cabal in which the virus is no more dangerous than a mild flu or the common cold; and (iv) the use of the pandemic as a covert operation supported by Bill Gates to develop a global surveillance regime facilitated by widespread vaccination. As the conversations evolve, these conspiracy theories appear to be connecting to one another, and may eventually form a single coherent conspiracy theory that encompasses all of these actants. At the same time, smaller nucleations of emerging conspiracy theories can be seen in the overall social media narrative framework graph. 

Because the news cycle appears to chase social media conversations, before feeding back into it, there is a pressing need for systems that can help to monitor the emergence of conspiracy theories as well as rumors that might presage real-world action. Already we have seen people damage 5G infrastructure, assault people of Asian heritage, deliberately violate public health directives, and ingest home remedies, all in reaction to the various conspiracy theories active in social media and the news. We have shown that a pipeline of interlocking computational methods, based on sound narrative theory, can provide a clear overview of the underlying generative frameworks for these narratives. Recognizing the structure of these narratives as they emerge on social media can assist not only in fact checking but also in averting potentially catastrophic actions. Deployed properly, these methods may also be able to help counteract various dangerously fictitious narratives from gaining a foothold in social media and the news. At the very least, our methods can help identify the emergence and connection of these complex, totalizing narratives that have, in the past, led to profoundly destructive actions.

\bibliographystyle{unsrt}  


\end{document}